\documentclass[journal, draftcls, one column, 12 pt]{IEEEtran}

\usepackage{graphicx}
\usepackage{epstopdf}
\usepackage{url}
\usepackage{braket}

\usepackage{graphicx}
\usepackage{wrapfig}
\usepackage{booktabs}       
\usepackage{amsfonts}       
\usepackage{nicefrac}       
\usepackage[numbers]{natbib}
\usepackage{color,multirow, multicol}
\usepackage{amsthm}
\usepackage{amsmath}
\usepackage{algorithm}
\usepackage[noend]{algpseudocode}
\DeclareMathOperator*{\argmin}{argmin} 
\theoremstyle{definition}

\newtheorem{lemma}{Lemma}

\newtheorem{proposition}{Proposition}
\usepackage{caption}
\usepackage{subcaption}
\newcommand{\mX}{{\mathcal X}}
\newcommand{\mY}{{\mathcal Y}}
\newcommand{\mZ}{{\mathcal Z}}
\newcommand{\mP}{{\mathcal P}}
\newcommand{\mD}{{\mathcal D}}

\usepackage{mathtools}

\DeclarePairedDelimiter\floor{\lfloor}{\rfloor}

\ifCLASSINFOpdf
\else
\fi
\begin{document}
%
\title{Optimal Piecewise Local-Linear Approximations }
%
%
%

\author{Kartik~Ahuja, 
	William~Zame,
	and~Mihaela~van der Schaar,~\IEEEmembership{Fellow,~IEEE}
	\thanks{K. Ahuja and Mihaela van der Schaar are with the Department
		of Electrical and Computer Engineering, University of California, Los Angeles,
		CA, 90095 USA e-mail: (see ahujak@ucla.edu, mihaela@ee.ucla.edu)}
	\thanks{William R. Zame is with Economics Department, University of California, Los Angeles}
}

%
%

\markboth{Journal of \LaTeX\ Class Files,~Vol.~14, No.~8, August~2015}%
{Shell \MakeLowercase{\textit{et al.}}: Bare Demo of IEEEtran.cls for IEEE Journals}
%



\maketitle

\begin{abstract}
	Existing works on ``black-box'' model interpretation use  local-linear approximations to explain the predictions made for each data instance in terms of the importance assigned to the different features for arriving at the prediction. These works provide instancewise explanations and thus give a local view of the model. To be able to trust the model it is important to understand the global model behavior and there are relatively fewer works which do the same. Piecewise local-linear models  provide a natural way to extend local-linear models to explain the global behavior of the model. In this work, we provide a dynamic programming based framework to obtain piecewise approximations of the black-box model. We also  provide provable fidelity, i.e., how well the explanations reflect the black-box model, guarantees. We carry out simulations on synthetic and real datasets to show the utility of the proposed approach.  At the end, we show that the ideas developed for our framework can also be used to address the problem of clustering for one-dimensional data. We give a polynomial time algorithm and prove that it achieves optimal clustering.
\end{abstract}



%
\IEEEpeerreviewmaketitle

\section{Introduction}

\IEEEPARstart{M}{achine}  learning algorithms have proved hugely successful for a wide variety of supervised learning problems.  However, in some domains, adoption of these algorithms has been hindered because the ``black-box'' nature of these algorithms makes their predictions  difficult for  potential users to interpret.  This issue is especially important  in the medical domain and security applications \cite{caruana2015intelligible}. The European Union's Law on Data Regulation  \cite{goodman2016european} makes it mandatory for ``black-box'' models to explain how they arrive at the predictions before implementing them in practice.

The problem of interpretation has received substantial attention in the literature recently \cite{ribeiro2016should} \cite{shrikumar2017learning} \cite{bastani2017interpreting} \cite{chen2018learning}.  These papers have approached the problem of  interpreting the black-box model by approximating it with local models (e.g., linear models) in a neighborhood of each data point (See the justification in \cite{lundberg2017unified}). \footnote{ We define piecewise models later.}  These papers provide instancewise explanations of the predictions made by the model and they only explain the local nature of the model. These papers do not provide insights into the global behavior of the model.   There are relatively fewer works \cite{bastani2017interpreting} \cite{lakkaraju2017interpretable} that explain the global model behavior. While linear models are common to use as local models, there are very few global explanation frameworks  that partition the feature space into \textit{homogeneous} regions and fit a local-linear model to predict the black-box function behavior. In  this paper, we provide a framework to build such piecewise local-linear approximations of the model. 

\subsection{Contribution} We are given a set of data points and the black-box function values computed for those data points. Our goal is to construct a partition of the feature space into subsets and to assign a simple local model to each subset.   We propose a  dynamic programming based approach to find the partition of the dataset and the set of local models.  We prove that the output of our method, which we refer to as a piecewise local-linear interpreter (PLLI), is approximately optimal in several different cases, i.e. it probably approximately correct  (PAC) learns the black-box function.   We use several real and synthetic datasets to establish the utility of the proposed approach. The code for the proposed method and experiments can be found at \url{https://github.com/ahujak/Piecewise-Local-Linear-Model}.

Our framework is very general and helps model interpretation in different ways described below.
\begin{itemize}

	\item \textbf{Region-wise feature importances:} We broadly categorize the works on feature importance scoring into two categories: a) Global feature importance scoring: Methods such as importance scoring based on tree based models \cite{archer2008empirical} fall in this category. These methods identify the factors that the model finds important overall when making the predictions across all the data points. b) Instancewise feature importance scoring: Methods such as \cite{chen2018learning} fall in this category. These methods identify the factors that the model finds important when making predictions for a certain data instance. Instancewise methods provide a more refined understanding of the model at an instance level while the global methods give an aggregate understanding of the model.  Using an instance based approach across all the data points to achieve global explanations is impractical. Therefore, it is important to find a middle ground between the global and instance based approaches. Our method helps achieve this task. We divide the feature space into regions and fit local-linear models and assign importance scores to different features in the different regions.

	
	%
	
	
	\item \textbf{Global explanations through instancewise explanations:} Many works on instancewise explanations try to provide global explanation to model behavior in terms of the best $K$ representative data instances and constructing explanations for them. 
	In \cite{ribeiro2016should} authors proposed a method to identify such representative data instances. Our framework divides the feature space into different regions and thus it naturally provides a method to identify different representative data points. We further elaborate on the utility of our approach in comparison to the method in \cite{ribeiro2016should} in the Experiments Section.
	
	\item At the end, we leverage ideas from our framework to show that our algorithm can be applied to the problem of clustering. We prove that our algorithm achieves optimal clustering for one-dimensional data in polynomial time. To the best of our knowledge, this is the first proof that a method can achieve optimal clustering for one-dimensional data.
	
\end{itemize}

%

\section{Problem Formulation}
\subsection{Toy Example}
\label{toy_example}
\begin{figure}[ht]
	
	\begin{center}
		\includegraphics[trim= 0mm 0mm 0mm 0mm,width=4in]{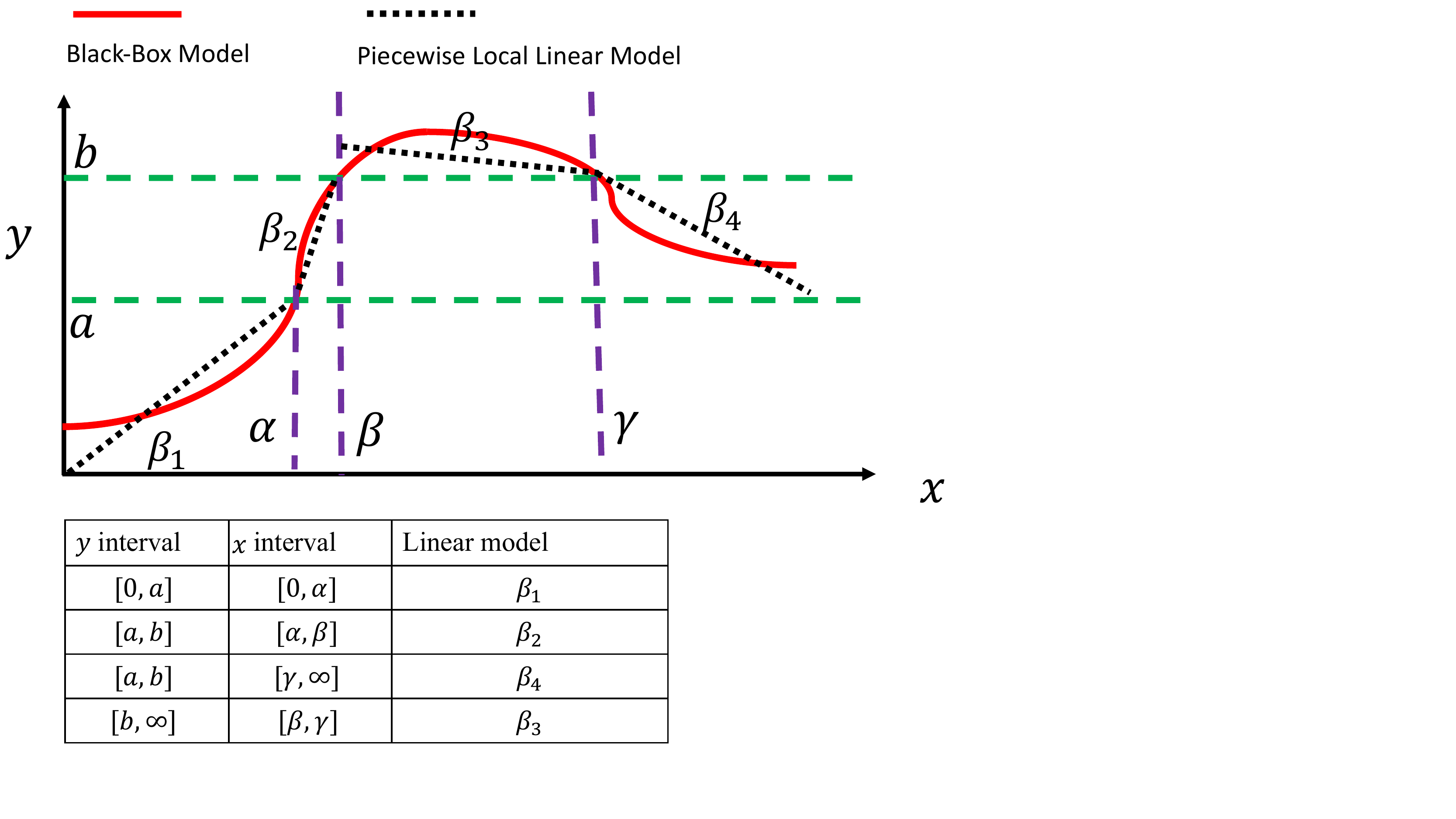}
		\caption{Comparison of the black-box model versus the piecewise approximation.}
		\label{figure_motivaton1}
	\end{center}
\end{figure}

In this section, we begin by describing a toy example to illustrate the input and the output from the proposed algorithm. In Figure \ref{figure_motivaton1}, we show the example of a one dimensional black-box function $f$. We input the data $D = \{(x_i, f(x_i))\}_{i=1}^{n}$ to the proposed algorithm, where $(x_i, f(x_i)) \in \mathcal{V}$, where $\mathcal{V}$ is the joint feature value and function value space.  Our goal is to partition the space $\mathcal{V}$ into homogeneous regions -- similar in terms of black-box function values and similar in terms of their features. We first partition the function $f$'s range into three intervals as shown in the figure (later we explain the methodology used to decide the intervals).

Next we want to partition the data that belongs to the inverse mapping of these intervals.  Consider the interval $[a,b]$ and its inverse  $f^{-1}[a,b] = \{[\alpha, \beta]\} \cup \{[\gamma, \infty)\}$. The inverse map $f^{-1}[a,b]$ is not a connected set.  Hence, it is natural to partition $f^{-1}[a,b]$ into two separate connected regions $[\alpha, \beta]$ and $[\gamma, \infty]$ (a natural approach to partition the data in $f^{-1}[a,b]$ is to use k-means clustering with  number of clusters $k=2$).  For general functions, also $f^{-1}[a,b]$ will not be connected and thus it is natural to partition the data in this region using k-means clustering. The choice of number of clusters  depends on number of disconnected sets, which is not known apriori. Hence, a natural approach is to select the select number of clusters using cross-validation.
The Table in Figure \ref{figure_motivaton1} summarizes the partition in terms of the $y$ interval and $x$ interval and the coefficients for local-linear models and  the black dotted line shows the piecewise local-linear approximation.

\begin{figure}[ht]
	
	\begin{center}
		\includegraphics[trim= 0mm 0mm 0mm 0mm,width=4in]{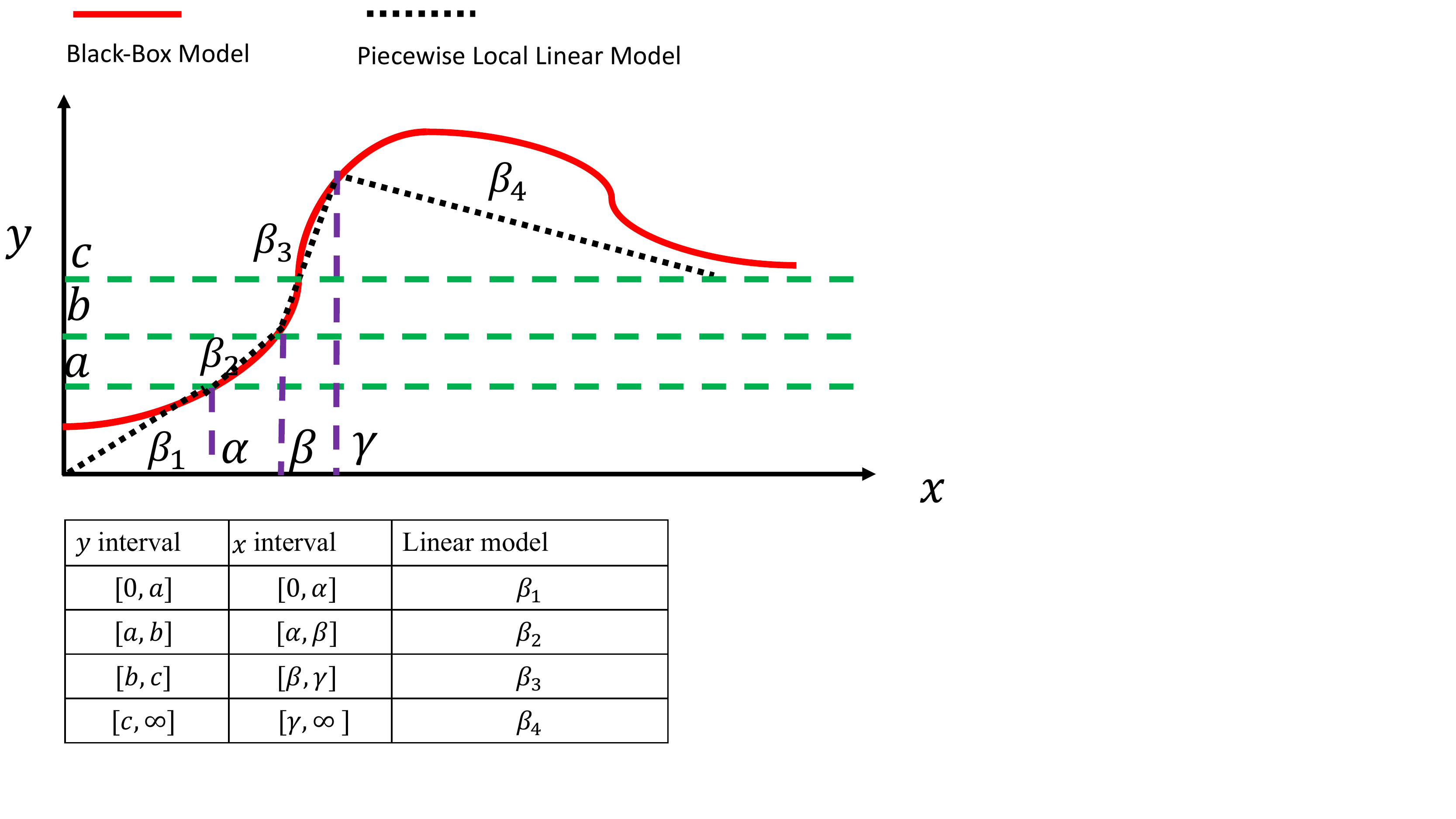}
		\caption{Comparison of the black-box model versus the piecewise approximation.}
		\label{figure_motivaton2}
	\end{center}
\end{figure}

\subsection{Interpretive models} 

We are given a space $\mX$ of {\em features} and a space  $\mY = [0,1]$ of {\em labels}.  We are given a  predictive model $f : \mX \to \mY$ (say a random forest based model or a deep neural network model). The data is distributed according to some {\em true distribution} 
$\mD$ (typically  unknown) on $\mX$.\footnote{We assume that the cumulative distribution function of $f$ is a  continuous function.}  Our objective is to interpret $f$ in terms of {\em interpretive models}, which  are defined  below.   We seek to find a interpretive model $g$ that approximates $f$. 

The intepretive models we consider here represent the most commonly used models in literature on model interpretation \cite{ribeiro2016should}\cite{lundberg2017unified}. The interpretive models we consider are defined by  partitioning $\mX$ into a finite number of disjoint sets and assigning a simple model (linear or constant model) to each set of the partition. 


To make this precise, recall that a (finite) partition of a subset $A \subset \mX$ is a family $\mZ = \{Z_1, Z_2, \ldots, Z_K\}$ of subsets of $\mathcal X$ such that $\bigcup_{i=1}^K Z_i = A$ and  $Z_i \cap Z_j = \emptyset$ if $i \not= j$.  
%
Given a partition 
${\mathcal Z}$ of $A$ and an instance $a \in A$ we write ${\mathcal Z}(a)$ for the index of the unique element of the partition ${\mathcal Z}$ to which $a$ belongs.  Write $\mP(A)$ for the set of all (finite) partitions of $A$ and $\mP_K(A)$ for the set of partitions having $K$ elements. We define $\mathcal{M} = \{M_1,...,M_K\}$ a set of local models, where model $M_j$ corresponds to the local model for points in $Z_j$. Each local model $M_{j}:\mathcal{X}\rightarrow \mathcal{Y}$ belongs to a set $\mathcal{H}$ of models, where $\mathcal{H}$ can be  from the family of  linear models ($M_j(x) = \max\{\min\{b^tx + c,1\},0\}$, where $b \in \mathbb{R}^{ |\mathcal{X}|}$ and  $c \in \mathbb{R}$), constant models ($M_j(x) = c$, where $c \in \mathbb{R}$). Given a partition $\mZ$ of $\mX$ and the corresponding set of local models $\mathcal{M}$, we define a {\em interpretive model} $g_{\mathcal{M}, \mathcal{Z}}: \mathcal{X} \to Y$  by $g_{\mathcal{M}, \mathcal{Z}}(x) = M_{\mZ(x)}(x)$.

\begin{figure}[ht]
	
	\begin{center}
		\includegraphics[trim= 0mm 70mm 0mm 0mm,width=4.25in]{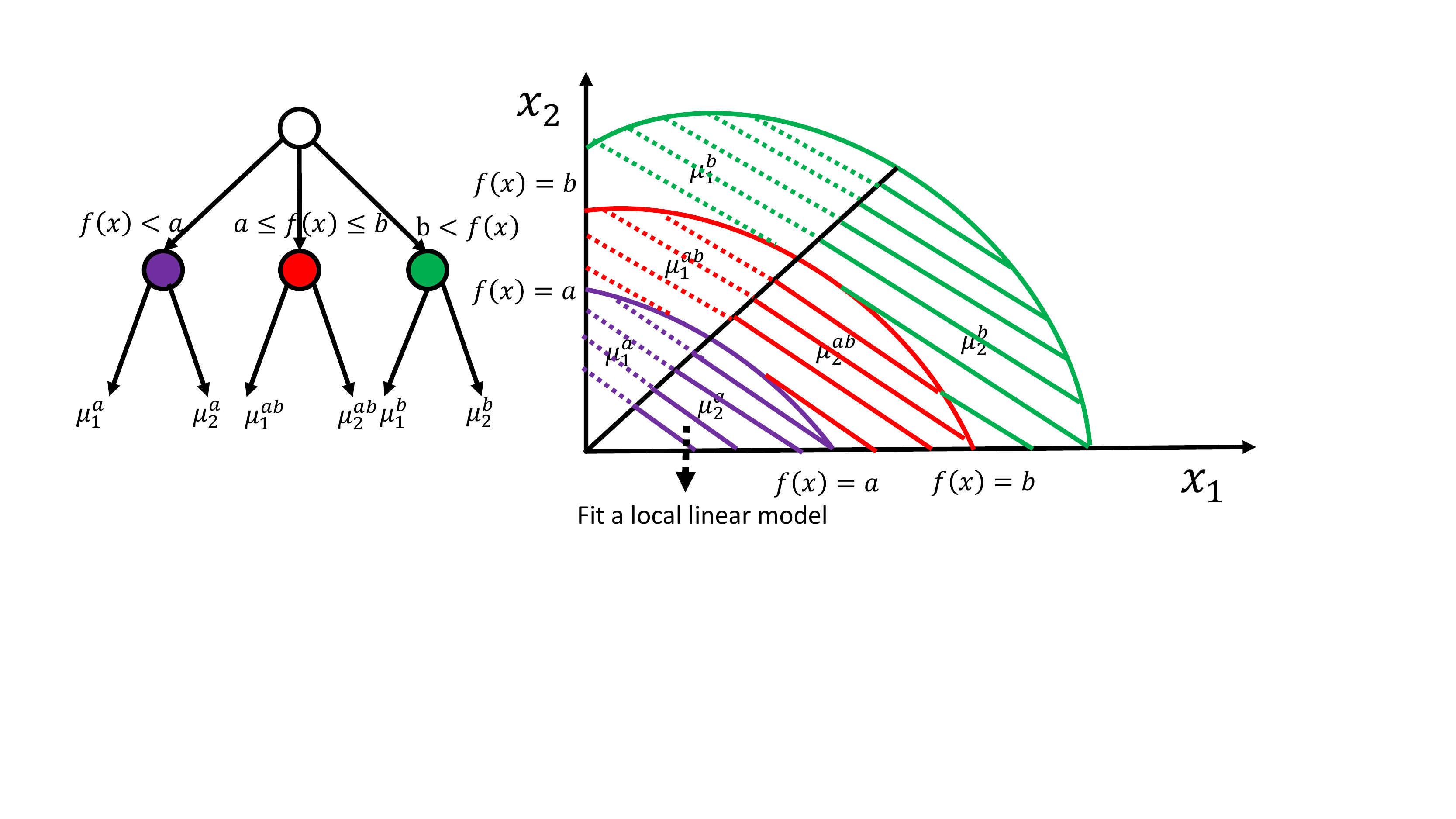}
		\caption{Example of a 2-D partition}
		\label{figure_partition_example}
	\end{center}
\end{figure}

We define the search space of our partitions next and note that we will closely follow the type of partitions described in the Toy example in Section \ref{toy_example}.  Partition the the range of function $f$ into $H$ intervals given as $\{a_r\}_{r=1}^{H-1}$, where $a_1<a_2,...<a_{H-1}$. Next we partition the data in the inverse mapping of these intervals, i.e., $f^{-1}[a_r, a_{r+1}] \in \mathcal{X}$, into $S_r$ regions. There can be many methods to partition the data into $S_r$ regions such as k-means clustering, hierarchical clustering etc.. The choice of the clustering algorithm is a hyperparameter and our method allows us to choose any of these clustering algorithms. However, in this work, for ease of exposition, we would only use k-means clustering algorithms \footnote{We assume that we use the same initialization or same seed for initialization in k-means. This is done to ensure  if the algorithm is used multiple times on the same data it should lead to the same outcome.} to partition the inverse images of the intervals into different regions.   We characterize region $i$ among these $S_r$ regions  by its centroid $\mu_i^r$, where the centroid is defined in as the mean of the data points in the region. All the points in region $i$ are closer to the centroid $\mu_i^r$ than to the centroids of the rest of the regions. A region $Z_k$ of the partition $\mathcal{Z}$ is defined as  $Z_k = \{x,\;\text{s.t.} \; x\in f^{-1}[a_r, a_{r+1}], \;\|x-\mu^{r}_i\| \leq \|x-\mu^{r}_j\|, \forall j\}$. We already showed a 1-D partition in Figure \ref{figure_motivaton1}, we give another example in Figure \ref{figure_partition_example} of a  2-D partition to illustrate more complicated shapes of the regions in the partition. The partition in Figure \ref{figure_partition_example} has six regions.

To summarize, we follow the approach of dividing the range of the function $f$ and then further dividing the inverse image of the $f$ into regions characterized by their centroids and we fit a linear model to explain the predictions in each region. The partition $\mathcal{Z}$ is characterized by the interval values $\{a_r\}_{r=1}^{H-1}, \{\{\mu_i^{r}\}_{i=1}^{S_r}\}_{r=1}^{H-1}$. The total number of regions in the partition is $\sum_{r=1}^{H-1}S_r$. 


\subsubsection{Why this type of partitions?}

 Our main purpose when constructing the partition is to find \textit{homogeneous} regions, i.e., regions where the data features $x$ are close to each other and the corresponding predictions $f(x)$ are also close to each other.  To achieve the first task, i.e., the function values are close, we first partition the range of function $f$.  However, just partitioning the range of $f$ does not guarantee that the inverse image of the intervals consists of data instances that are also close to each other. See $f^{-1}[a,b]$ in the Toy example in Figure \ref{figure_motivaton1}, it consists of two disconnected regions. This is the reason why we  partition the inverse image of each interval $f^{-1}[a_r, a_{r+1}]$ such that any disconnected regions are separated into different regions.


\subsubsection{Comment on the choice of the number of intervals and number of regions}
How many intervals $H$ should we divide the function's range into? How many regions should we further subdivide those intervals into $S_r$? In Figure \ref{figure_motivaton1} and Figure \ref{figure_motivaton2}, we divide the feature space $\mathcal{X}$ into four regions with different choices of $H$ and $\{S_r\}_{r=1}^{H}$. We compare and select between the two choices using cross-validation. In general, if we want to construct $K$ regions there can be many possible choices for $H$ and $\{S_r\}_{r=1}^{H}$, where $\sum_{r=1}^{H}S_r = K$. In this work, we will carry out simulations assuming that $S_r = S_t=W$ for any $r\not=t$. The more general case when    $S_r \not= S_t$ will be a part of future work.

Before we describe the main algorithm, we fix some hyperparameters. We assume that $K$ is given (provided as input by the expert or a parameter that can be tuned using cross-validation).
Suppose that we want to have $K$ regions in the partition. We assume that we will divide the function's range space into $H$ intervals and each interval into $W$ regions. Therefore, $H\times W = K$.\footnote{The maximum number of choices for $(H,W)$ are $K$.}
 Hence, the partition $\mathcal{Z}$ is characterized by the interval values $\{a_r\}_{r=1}^{H-1}, \{\{\mu_i^{r}\}_{i=1}^{W}\}_{r=1}^{H-1}$ where each region  $Z_k$ in $\mathcal{Z}$ is  $Z_k = \{x,\;\text{s.t.} \; x\in f^{-1}[a_r, a_{r+1}], \;\|x-\mu^{r}_i\| \leq \|x-\mu^{r}_j\|, \forall j\}$.  The set of all such partitions $\mathcal{Z}$ is written as $\mathcal{P}_{K}(\mathcal{X})^{\dagger}$. Recall here we are assuming that each interval is partitioned $W$ regions using k-means clustering algorithm.



\subsection{Loss functions} We measure the fidelity \cite{tan2018distill} of a proposed interpretation $g$ for $f$ in terms of a given {\em loss function} defined as $\ell : \mathbb{R}_{+} \to  {\mathbb R}_+$.  
We define the risk/fidelity achieved by model $g_{\mathcal{M}, \mathcal{Z}}$ as follows
\begin{equation}
R(f,g_{\mathcal{M}, \mathcal{Z}}; \mathcal{D}) = E_{\mathcal{D}}[\ell( |f(X)  -g_{\mathcal{M}, \mathcal{Z}}(X)|)]
\end{equation}
where $X$ is a feature from the distribution $\mathcal{D}$, the  expectation is taken over the distribution $\mathcal{D}$.  If $\ell$ is square function, then  we obtain mean square error (MSE). 

\subsection{Expected Risk Minimization}
Our objective is to find a partition 
${\mathcal Z}$ and corresponding set of local models $\mathcal{M}$ (where each local model is drawn from $\mathcal{H}$) to minimize the true risk subject to the constraint that the size of the partition, i.e., $|\mZ|=K$ . 
\begin{equation}
(\mathcal{M}^*, \mZ^*) = \argmin_{\mathcal{M}\in \mathcal{H}^{K}, \mZ \in \mP_K(\mX)^{\dagger}} R(f, g_{\mathcal{M}, \mZ}; \mD)  \label{eqn:truerisk} 
\end{equation}

$g_{\mathcal{M}^{*}, \mZ^{*}}$ is the best piecewise model that minimizes the  above risk. 


\subsection{Empirical Risk Minimization } 
In practice, we do not know the true distribution $\mathcal{D}$ so we cannot minimize the true risk. We are given a finite dataset (training set) 
$D = \{(x_i, f(x_i))\}_{i=1}^{N}$ drawn from the true distribution as input.  For given $\mathcal{M}, \mZ$ the empirical risk is

\begin{equation}
\hat{R}(\mathcal{M}, \mathcal{Z}; D) =  \frac{1}{n} \sum_{(x_i, f(x_i)) \in D} \ell(|f(x_i)- g_{\mathcal{M}, \mZ}(x_i)|) 
\label{emp-risk}
\end{equation}

The spirit of Probably Approximately Correct (PAC) learning \cite{shalev2014understanding} suggests that we should minimize the empirical risk:
\begin{equation}
(\mathcal{M}^{\dagger}, \mZ^{\dagger}) = \argmin_{\mathcal{M} \in \mathcal{H}^{K},  \mZ \in \mP_K(\mX)^{\dagger}}  \hat{R}(\mathcal{M}, \mathcal{Z}; D) 
\label{minim-prob}
\end{equation}

Later we will show that solving the above empirical risk minimization problem is the PAC solution to the actual risk minimization problem in \eqref{eqn:truerisk}. We cannot solve the above problem using brute force search because it requires searching among $\mathcal{O}(|D|^K)$ partitions, which becomes intractable very quickly with increase in $|D|$ and $K$.  In the next section, we propose an efficient algorithm to solve \eqref{minim-prob}.

\section{Piecewise Local-Linear Interpreter }

In this section, we develop the Piecewise Local-Linear Interpreter (PLLI) to solve the problem discussed above.  Without loss of generality we assume that all the data points $x_i$ in $D$ are sorted in the increasing order  of $f(x_i)$.


There are two parts to the Algorithm (Algorithm 1 and 2). 
In the first part, the Algorithm partitions the data $D$ into subsets and finds an optimal local model corresponding to each subset.  The division of the dataset into these subsets relies on dynamic programming.     The risk achieved by partition of first $m$ points into $w$ intervals is defined as $V^{'}(m,w)$, where the intervals are meant to divide the range of the function $f$.

For each  $x_i , x_j$ in the dataset, where $i \leq j$, define a subset of the data  as follows.

$D(i,j) = \{x : x\in D  \; \text{\&}\;  f(x_i) \leq f(x) < f(x_j)\}$

The dataset $D(i,j)$ is to be partitioned in different regions  of the feature space. We divide the dataset $D(i,j)$ into $W$ regions using k-means clustering \footnote{We can use other clustering methods as well instead.} with $k=W$, where the regions are given as $\{S_{1}^{ij},...,S_{W}^{ij}\}$. We fit a linear model separately to each of these regions. We define the total risk achieved over the dataset $D(i,j)$ by these $W$ local models as  $G(i,j)$ in \eqref{eqn: centroid}.


\begin{figure*}
	\begin{equation}
	\begin{split}
	G(i,j) =\sum_{u}\min_{h \in \mathcal{H}} \sum_{x_r \in S^{ij}_{u}} \ell(|f(x_r)-h(x_r)|) \bigr]   \\ 
	\label{eqn: centroid}
	\end{split}
	\end{equation}
\end{figure*}
Suppose  the Algorithm wants to divide the first $p$ points into $q$ intervals. Also, suppose that the Algorithm has already constructed a partition to divide the first $m$ points into $w$ intervals for all $m\leq p-1$ and for all $w\leq q-1$. The Algorithm divides $p$ points into $q$ intervals as follows
\begin{equation*}
\begin{split}
V^{'}(p,q) = \min_{n^{'}\in \{1,..,p-1\}} \big[V^{'}(n^{'},q-1) + G(n^{'}+1, p)\big]  \\
\Phi(p,q) = \argmin_{n^{'}\in \{1,.,p-1\}} \big[V^{'}(n^{'},q-1) + G(n^{'}+1, p)\big]
\end{split} 
\end{equation*}

where $\Phi(p,q)$ is the index of the first data point in the $q^{th}$ interval. The interval $q$ consists of all the points  indexed $\{\Phi(p,q),..p\}$.  The data subset defined as $D(\Phi(p,q),p)$ and it is partitioned into $W$ subsets each fitted with its own local model. Similarly, the next subset, i.e., the $(q-1)^{th}$ interval can be computed recursively as $\{\Phi(\Phi(p,q), q-1),..,\Phi(p,q)-1\}$ and so on. 

 In the first part of the Algorithm, we construct a partition of $D$ and the corresponding set of local models.  In the second part of the Algorithm,  we extend this partition from the dataset $D$ to the set  $\mathcal{X}$. We write the function that is output by the Algorithm 2 as $g_{M^{\#}, \mathcal{Z}^{\#}}$.

\begin{algorithm*}[tbh]
	\caption{ Computing value and index functions}
	\label{alg:example}
	\begin{algorithmic}[1]
		\State \textbf{Input:} Dataset $D$, Number of intervals $H$ and number of regions to divide each interval $W$
		\State\textbf{Initialize:} Define $V^{'}(1,k)=0, \forall k \in \{1,...,K\}$. 
		\State  For each  $x_i \in D, x_j \in D$ such that $i \leq j$, define $D(i,j) = \{x : x\in D  \; \text{and}\; f(x_i) \leq f(x) \leq f(x_j)\}$
		\State $ \{S_{1}^{ij},..,S_{W}^{ij}\}= \mathsf{Kmeans}(D(i, j))$
		\State  $G(i,j) = \sum_{u}\min_{h \in \mathcal{H}} \sum_{ x_r \in S^{ij}_{u}} \ell(|f(x_r) -h(x_r)|)  $ 
		\State  $M^{'}(S_{u}^{ij}) = \arg\min_{h \in \mathcal{H}} \sum_{ x_r \in S^{ij}_{u}} \ell(|f(x_r) -h(x_r)|)  $ 		
		\For{ $n \in \{2,..., |D|\}$}
		\For{$k \in \{1,..., H\}$}
		\State $\;\;\;\;\;\;\;\;\;\;$ \begin{equation} V^{'}(n,k) = \min_{n^{'}\in \{1,..,n-1\}} \big[V^{'}(n^{'},k-1) + G(n^{'}+1, n)\big] \end{equation} 
		\State  $\;\;\;\;\;\;\;\;\;\;$ \begin{equation} \Phi(n,k) = \argmin_{n^{'}\in \{1,..,n-1\}} \big[V^{'}(n^{'},k-1) + G(n^{'}+1, n)\big] \end{equation}
		\EndFor
		\EndFor
		\State  \textbf{Output:} Value function $V^{'}$, Index function $\Phi$			
	\end{algorithmic}
\end{algorithm*}

\begin{algorithm}
	\caption{ Computing partitions using the index function}
	\label{alg:algorithm2}
	\begin{algorithmic}[1]
		\State  \textbf{Input:} Index function $\Phi$, black-box predictive model $f$
		\State \textbf{Initialization:} $h_u =|D|, r=1$
		\For{ $k \in \{1,...,H\}$}
		\State  $h_l = \Phi(h_u,K-k+1)$ 
		\State $\{\mu^{k}_i\}_{i=1}^{W} =\mathsf{Kmeans}(D(h_l, h_u))$
		\For{$u \in \{1,...,W\}$}
		\State $\;$ $Z_{K-r+1} = \{x:f(x_{h_l})<f(x)\leq f(x_{h_u}), $
		\State$\;\; \|x-\mu^{k}_{r}\| \leq \|x-\mu^{k}_{j}\|  \}$
		\State $\;$ $M_{K-r+1} = M^{'}(S^{h_l h_u}_{u} )$
		\State $\;$ $h_u = h_l$
		\State $\;$ $r=r+1$
		\State 
		\EndFor
		\EndFor
		\State  \textbf{Output:} $\mathcal{Z}^{\#} = \{Z_1,...,Z_K\}$, 
		\State $\mathcal{M}^{\#} = \{M_1,...,M_K\}$
	\end{algorithmic}
\end{algorithm}

\section{Main Results}
\label{Main_results}

Our goal in this section is to show that the output of the Algorithm PAC learns $f$. 


%
%


\subsection{PAC learnability of Algorithms 1 and 2:}

We state certain assumptions  before  the propositions.
\begin{itemize}
\item \textit{Assumption 1:} $\ell$ is a square function, i.e., $\ell(x) = x^2$. If $\ell$ is square function, then the objective in \eqref{eqn:truerisk}, \eqref{minim-prob}  is MSE.


\item \textit{Assumption 2:} The family of the local models $\mathcal{H}$ are linear models \footnote{We assume that the vector $\beta$ characterizing the linear function $\beta^tx $ has a bounded norm, i.e. $\|\beta\| \leq M$}

\item \textit{Assumption 3:} $H=K$ and $W=1$, we only partition the range of $f$ and do not further subdivide the intervals,

\end{itemize}
 In Propositions 1-4, we make Assumption 1. But following every proposition, we discuss how they generalize to other loss functions $\ell$.  
 To show PAC learnability, we will first show that the outcome of Algorithms 1 and 2 achieves the minimum empirical risk. 
\begin{proposition} \textit{Empirical Risk Minimization: } If the Assumption 1-2 hold, then
	the output of the Algorithm 1 and 2 achieves the minimum  risk value equal, i.e., $\hat{R}(\mathcal{M}^{\#},\mZ^{\#}; D) = \min_{\mathcal{M} \in \mathcal{H}^{K},  \mZ \in \mP_K(\mX)^{\dagger}} \hat{R}(\mathcal{M}, \mathcal{Z}; D)$.
\end{proposition}
The proof of Proposition 1 is given in the Appendix Section.  As you will see later, the proof of Proposition 1 does not rely on the fact that $\ell$ is square function. In fact the Proposition holds for any loss function $\ell$. In the above proposition, we showed that the Algorithm solves empirical risk minimization (ERM). In the next proposition, we show that the output of Algorithm 1 achieves minimization of the expected risk \eqref{eqn:truerisk}.

\begin{proposition} \textit{Expected Risk Minimization: }  If the Assumption 1-3 hold, then
	$\forall \epsilon>0, \delta\in (0,1), \exists\; n^{*}(\epsilon, \delta)$  such that if $D$ is drawn i.i.d. from $\mathcal{D}$ and $|D| \geq n^{*}(\epsilon, \delta)$ , then with probability  at least $1-\delta$,

	$|R(f,g_{\mathcal{M}^{\#}, \mathcal{Z}^{\#}}; \mathcal{D})-R(f,g_{\mathcal{M}^{*}, \mathcal{Z}^{*}}; \mathcal{D})|\leq \epsilon$
\end{proposition}

The proof of the Proposition 2 is in the Appendix Section.  In Proposition 2, we assume that $\ell$ is a square function. Proposition 2 can be generalized to the case when $\ell(|.|)$ is a Lipschitz continuous function.

\section{Experiments}

In this section, we describe the experiments conducted on synthetic and real datasets. We will cover regression problems in this experiments section.  The proposed method can also be applied to classification problems. All the simulations were conducted in Python in Google Colab. The code is available at  \url{https://github.com/ahujak/Piecewise-Local-Linear-Model}

\subsection{Metrics} We will use two metrics to measure the performance of the different methods.  We use MSE to measure the performance of the model on the labelled data (squared of the norm of the difference between the predictions and labels), which is denoted as MSE-p. We also use MSE to measure the fidelity  (squared of the norm of the difference between the predictions and black-box function values), i.e., how close is the model to the black-box model, which is denoted as MSE-f.  We use R$^2$, i.e., the coefficient of determination, to measure the fit of the model.

\subsection{Synthetic Dataset}
We begin by describing a synthetic dataset that we use to illustrate the performance of the method before going into a more complicated real data setting. We assume that each data point is of the form $(x_1,x_2,y)$, where $x_1$ and $x_2$ are the features and $y = (x_1+x_2)^2$. We assume that $x_1$ and $x_2$ are independent and are drawn from $\mathcal{N}(0,1)$. We sample 1000 data points $(x_1, x_2,y)$. 

\subsubsection{Black-Box Model} We split the data randomly into 80 percent training and 20 percent testing. We fit a random forest (RF) regressor to predict the target variable. In the Table \ref{table1: RF_reg_synthetic}, we compare the performance of the RF regressor, which is a black-box method, with other more interpretable methods such as a regression tree and a linear model. We observe that the RF regressor has a much smaller MSE-p in comparison to a linear model or a regression tree. We do not report R$^2$ for the linear model and regression tree as the models are so poor a fit that we obtained a negative R$^2$. Next, we will use PLLI to interpret this trained RF regression model.

\subsubsection{Piecewise Local-Linear Interpreter} 
\begin{figure}[!htb]
	\begin{center}
		
		\includegraphics[trim= 0mm 60mm 0mm 10mm,  width=4 in]{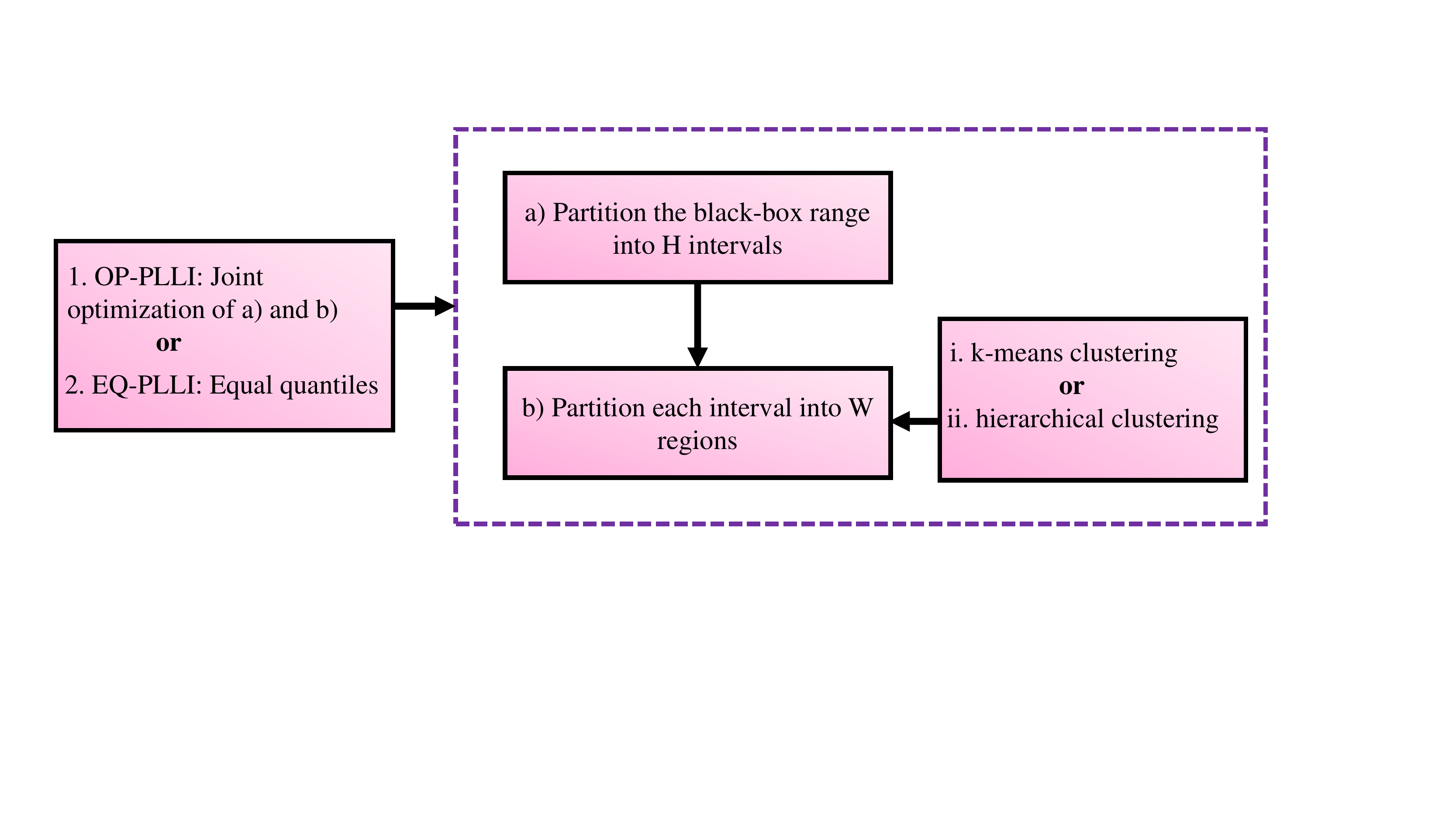}
		\caption{Overview of the PLLI framework}
		\label{fig:overview}
	\end{center}
\end{figure}

There are several possible configurations for the PLLI. We will fix $K=4$ (we fix a small value as the dataset is small).   
 We have three parameter configurations possible. ($H=4$, $W=1$), ($H=1$, $W=4$), ($H=2$, $W=2$). Instead of using the dynamic programming procedure described in Algorithm 1, we can alternatively use a simpler procedure to partition the function's range. We first order the dataset in terms of the black-box predictions and divide the dataset into $H$ equal quantiles. We use k-means clustering for data in each quantile to divide the data into $W$ clusters and fit a local-linear model to it. We refer to this procedure as EQ-PLLI, where EQ stands for equal quantile. On the other hand, we refer to the procedure from Algorithm 1 and 2 as OP-PLLI, where OP stands for optimal. In Figure \ref{fig:overview}, we show the overview of the PLLI framework with the various possible options in terms of the algorithms to construct the partition.

   In Table \ref{table2: RF_int_synthetic}, we compare these configurations in terms of MSE-f and MSE-p. Based on MSE-f and MSE-p, we select the OP-PLLI (H=2, W=2). In Table \ref{table2: RF_int_synthetic}, we also compare PLLI with other models. We fit a regression tree (with four leaves since $K=4$) based explanation model  from \cite{bastani2017interpreting} and a linear regression model  to predict the black-box model and find that these approaches do not perform well. 
\subsubsection{Black-Box Model Interpretation} 
\begin{figure*}[h]
	\begin{minipage}{.24\textwidth}
		\centering
		\includegraphics[width=1.8in]{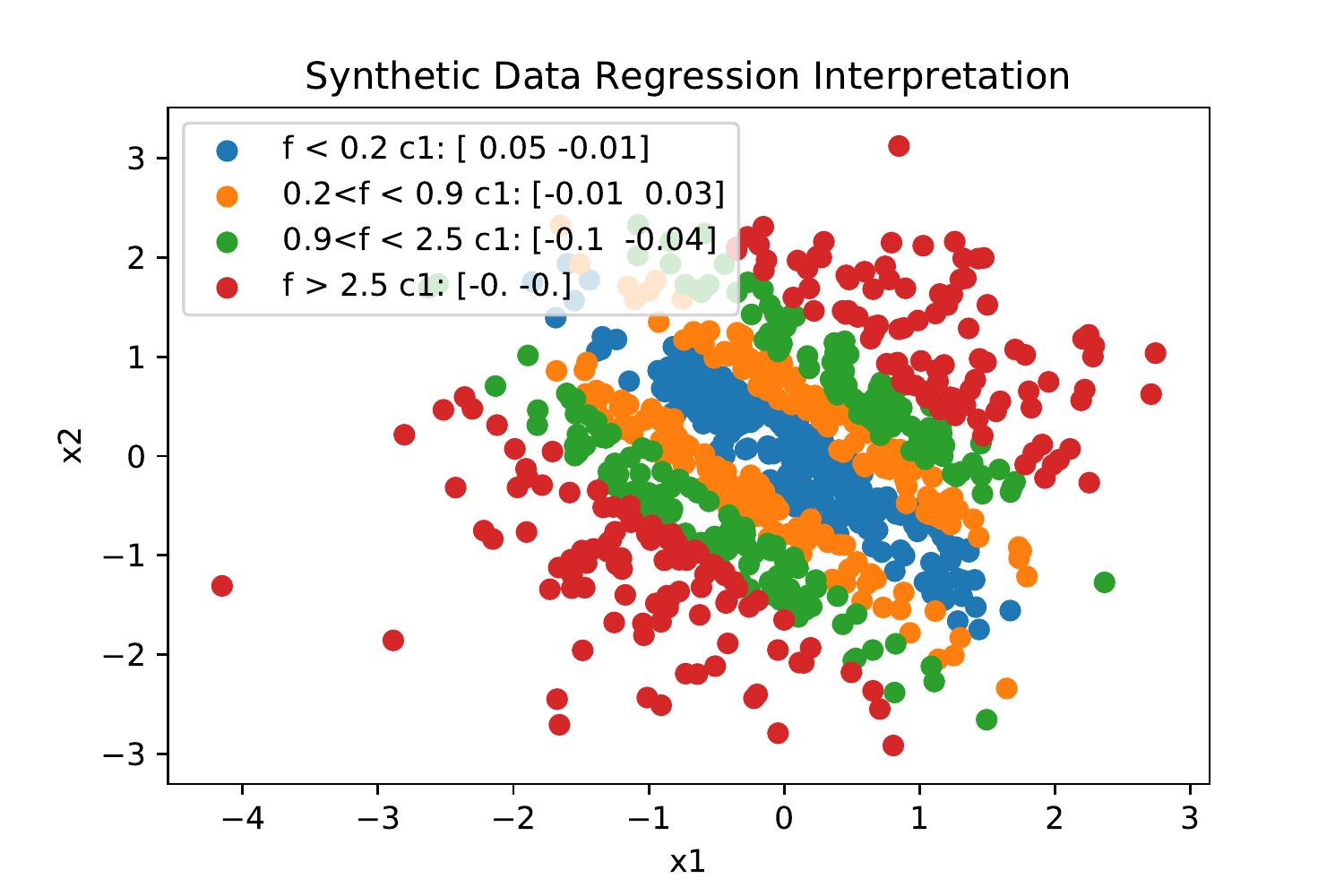}
		\label{fig:test1}
		\subcaption{H=4, W=1, EQ}
	\end{minipage}
	\begin{minipage}{.24\textwidth}
	\centering
	\includegraphics[width=1.8in]{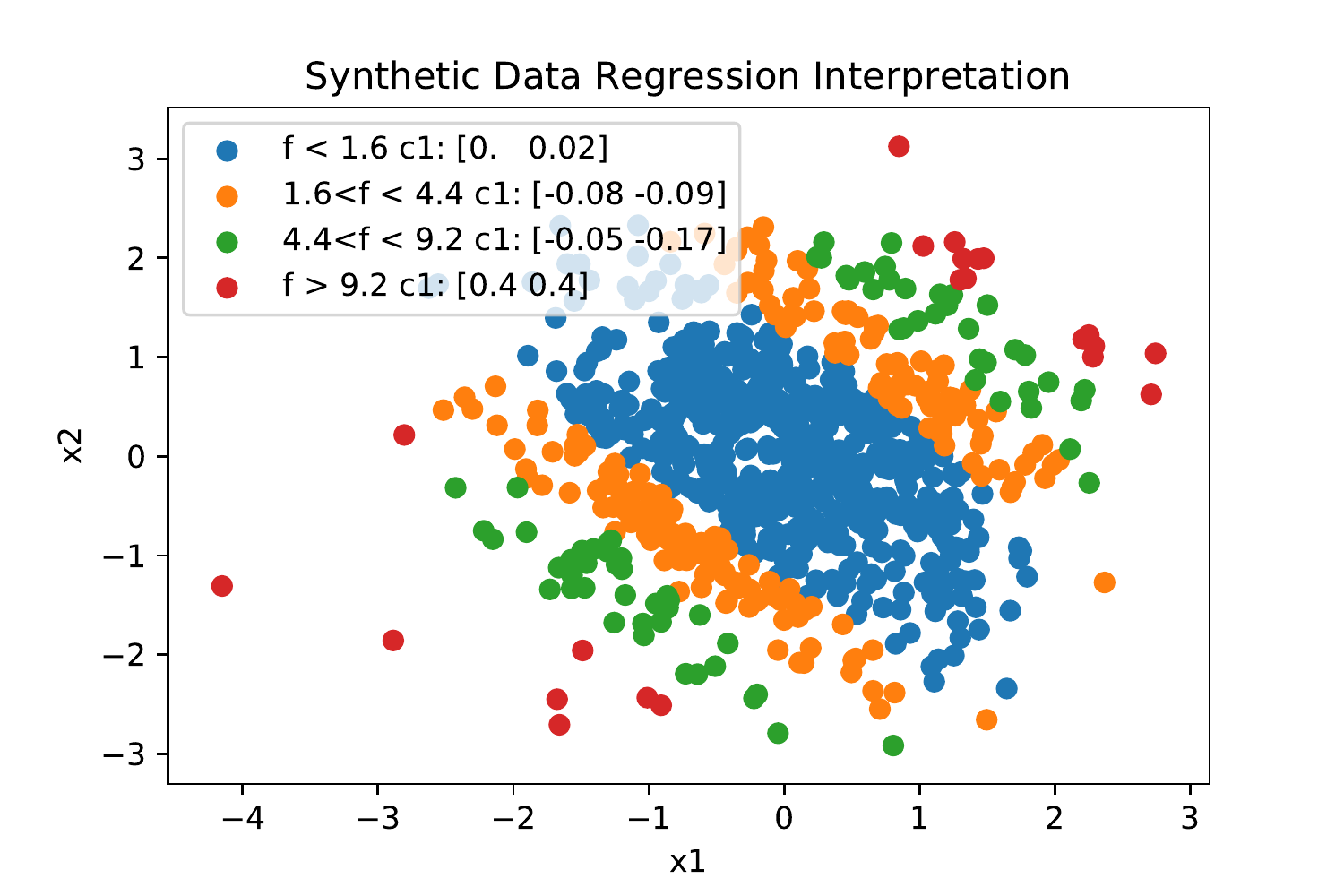}
	\label{fig:test4}
	\subcaption{H=4, W=1, OP}
\end{minipage}
	\begin{minipage}{.24\textwidth}
		\centering
		\includegraphics[width=1.8in]{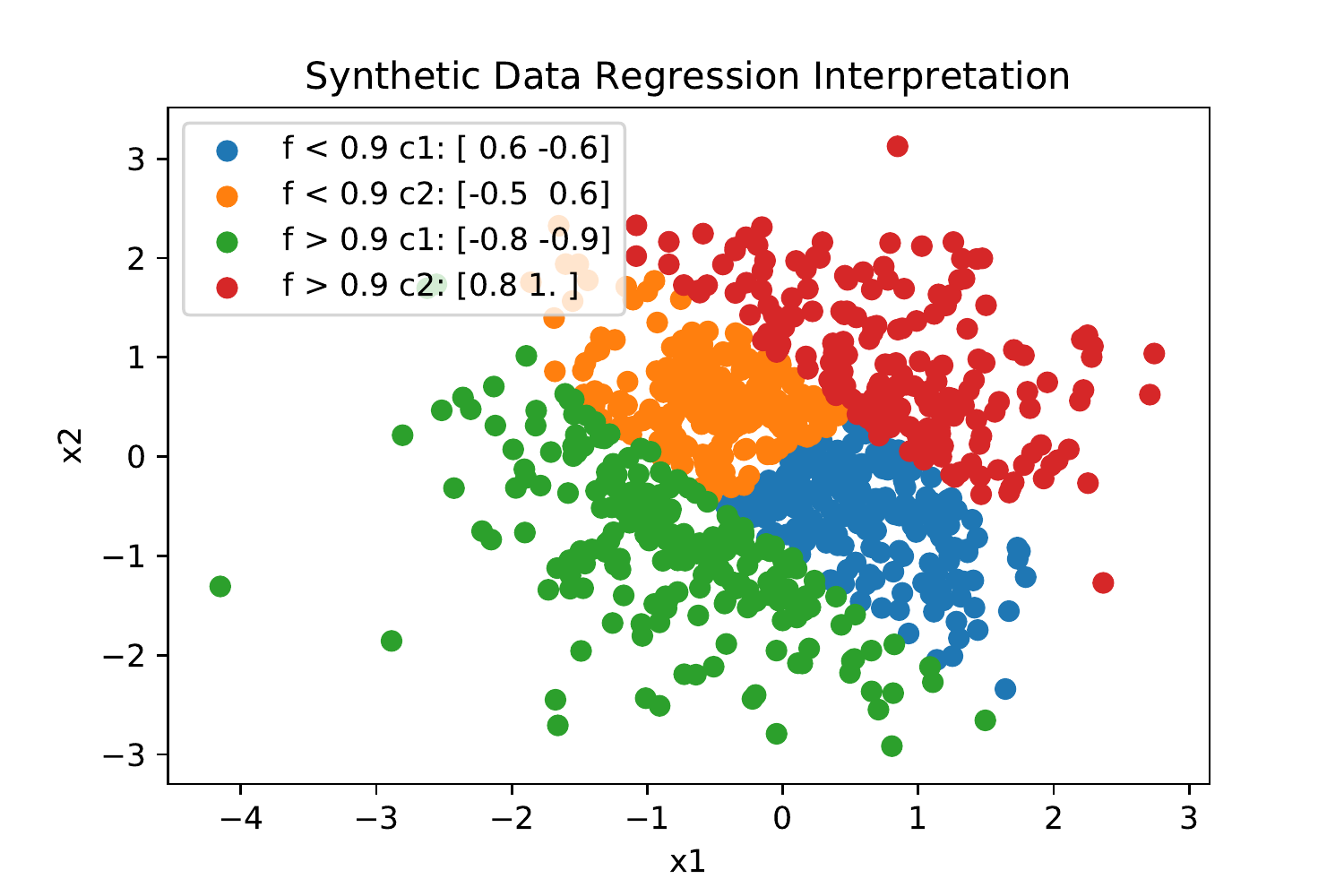}
		\label{fig:test2}
		\subcaption{H=2, W=2, EQ}
	\end{minipage}
	\begin{minipage}{.24\textwidth}
		\centering
		\includegraphics[width=1.8in]{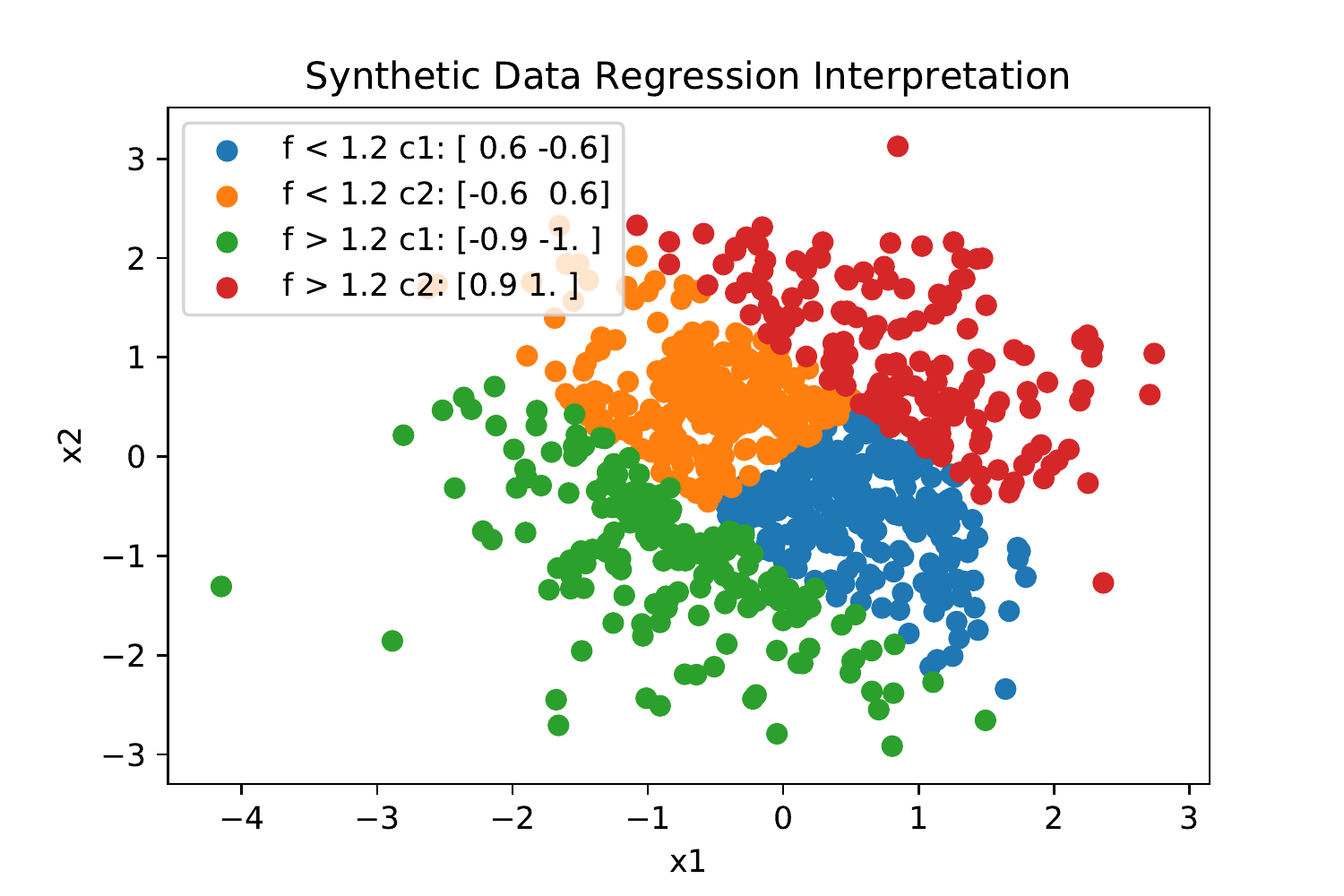}
		\label{fig:test3}
		\subcaption{H=2, W=2, OP}
	\end{minipage}

	\caption{Comparison of the piecewise model for different hyperparameter configurations}
	\label{fig1: RF_reg_synthetic}
\end{figure*}

\begin{figure}
	\begin{center}
		
		\includegraphics[trim= 0mm 0mm 0mm 0mm,  width=2 in]{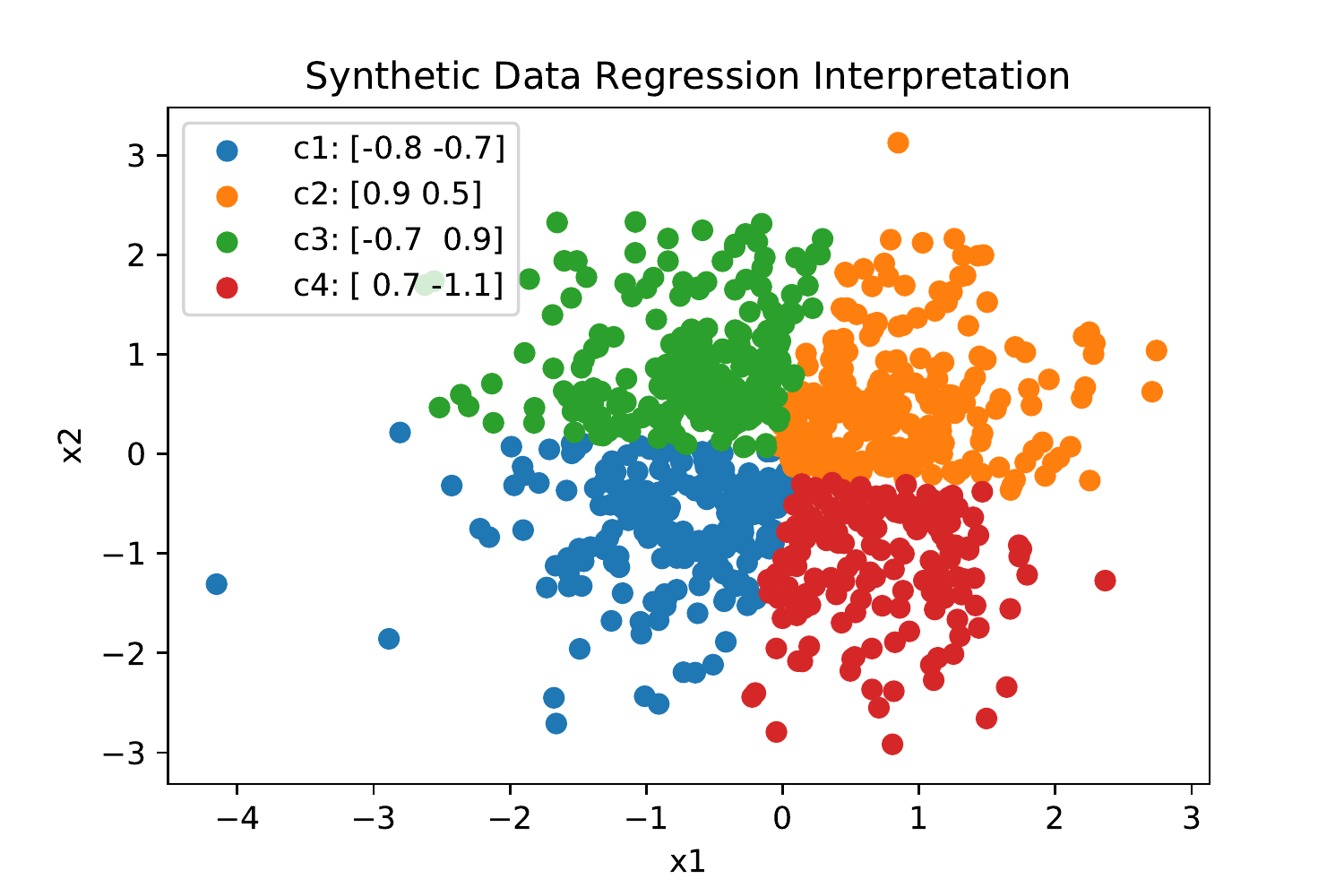}
		\caption{H=1, W=4, OP}
		\label{fig2: RF_reg_synthetic}
	\end{center}
\end{figure}

In Figures \ref{fig1: RF_reg_synthetic}, \ref{fig2: RF_reg_synthetic}, we show the partitions constructed under the different models shown in Table \ref{table2: RF_int_synthetic}. In Figure \ref{fig1: RF_reg_synthetic} a) and b) we show the different partitions from equal quantile and optimal approach. We  observe that the regions in each partition in Figure \ref{fig1: RF_reg_synthetic} a) and b) are not contiguous. For instance, the orange region is located in two separate areas of the 2-D feature space. In Figure \ref{fig1: RF_reg_synthetic} c) and d) the regions in each partition are contiguous.  In Table \ref{tabl3: model summary synthetic}, we give the region-wise feature importance constructed based on OP-PLLI with $(W=2, H=2)$. Each region is characterized by the centroid and the function range and the corresponding region-wise feature importance of x1 and x2 are shown in Table \ref{tabl3: model summary synthetic}.


\begin{table}[!h]
	\renewcommand{\arraystretch}{1.2}
	\centering
	\caption{Comparison of RF Regressor with other methods}
	\begin{tabular}[t]{|c|c|c|}
		\hline
		\textbf{Model} & \textbf{MSE-p} & \textbf{R}$^2$  \\ \hline
		RF Regressor & 0.11 & 0.97 \\ \hline 
		Regression tree & 6.30 & 0.00 \\ \hline 
		Linear model & 5.15 & 0.05\\ \hline 
		Constant model & 5.43 & --\\ \hline 
	\end{tabular}
\label{table1: RF_reg_synthetic}
\end{table}	

\begin{table}[!h]
	\renewcommand{\arraystretch}{1.2}
	\centering
	\caption{Comparison of PLLI (for various hyperparameter configurations) with other methods}
	\begin{tabular}[t]{|c|c|c|}
		\hline
		\textbf{Model} & \textbf{MSE-f} & \textbf{MSE-p}  \\ \hline
		EQ-PLLI (H=4, W=1) & 1.19 & 1.52\\ \hline 
		OP-PLLI (H=4, W=1) & 0.54& 0.73 \\ \hline 
		PLLI (H=1, W=4) & 0.69& 0.70 \\ \hline 
		EQ-PLLI (H=2, W=2) & 0.18 & 0.12\\ \hline 
		OP-PLLI (H=2, W=2) & 0.18& 0.11\\ \hline 
		Linear   model             & 5.75 & 7.57 \\ \hline
		Regression tree explainer \cite{bastani2017interpreting} &4.34 & 5.15\\ \hline
	\end{tabular}
\label{table2: RF_int_synthetic}
\end{table}	

\begin{table}[!h]
	\renewcommand{\arraystretch}{1.2}
	\centering
	\caption{Region-wise feature importance}
	\begin{tabular}[t]{|c|c|c|}
		\hline
		\textbf{Region} & \textbf{x1} & \textbf{x2}      \\ \hline
		$f<1.2, \; [0.6, -0.6]$ &0.104&0.002 \\ \hline		
		$f<1.2, \; [-0.6, 0.6]$ &0.070&0.004 \\ \hline		
		$f>1.2, \; [-0.9, -1.0]$ &4.230& 4.230 \\ \hline		
		$f>1.2, \; [0.9, 1.0]$ &4.230 & 4.230	\\ \hline		
	\end{tabular}
	\label{tabl3: model summary synthetic}
\end{table}	
\subsection{Interpret RF regression on Boston Housing Dataset}
We use the Boston Housing Dataset from UCI repository. 	The dataset consists of information about the house prices and other attributes about where the house is located. The attributes with their abbreviations and descriptions are described below. 
\begin{itemize}
	\item CRIM: per capita crime rate by town
	\item ZN: proportion of residential land zoned for lots over 25,000 sq.ft.
	\item INDS: proportion of non-retail business acres per town
	\item CHAS: Charles River dummy variable (= 1 if tract bounds river; 0 otherwise)
	\item NOX: nitric oxides concentration (parts per 10 million)
	\item RM: average number of rooms per dwelling
	\item AGE: proportion of owner-occupied units built prior to 1940
	\item DIS: weighted distances to five Boston employment centres
	\item RAD: index of accessibility to radial highways
	\item TAX: full-value property-tax rate per $10,000$
	\item PTR:  pupil-teacher ratio by town
	\item B 1000$(Bk - 0.63)^2$ where Bk is the proportion of blacks by town
	\item LST:   lower status of the population
\end{itemize}
The total number of instances in the dataset is 506.

\subsubsection{Black-Box Model}

We split the data randomly into 80 percent training and 20 percent testing. We fit a RF regressor to predict the target variable, i.e., the price of the house based on the attributes described above.  In the table below, we compare the MSE-p of the RF regression model and compare it with other interpretable methods such as a linear model and a regression tree model.  We observe that the RF regressor has a much smaller MSE-p in comparison to a linear model or a regression tree. However, the improvement in the performance comes at the cost that the new model is harder to interpret. In the next section, we use PLLI  to get insights into the behavior of this RF regressor  model. 
\begin{table}[!h]
	\renewcommand{\arraystretch}{1.2}
	\centering
	\caption{Comparison of RF Regressor with other methods}
	\begin{tabular}[t]{|c|c|c|}
		\hline
		\textbf{Model} & \textbf{MSE-p} & \textbf{R}$^2$  \\ \hline
		RF Regressor & 8.217& 0.88 \\ \hline 
		Regression tree & 37.23& 0.60 \\ \hline 
		Linear model & 21.78& 0.78 \\ \hline 
				Constant model & 81.54& 0.78 \\ \hline 
	\end{tabular}
\end{table}

\begin{table}[!h]
	\renewcommand{\arraystretch}{1.2}
	\centering
	\caption{Comparison of PLLI (for various hyperparameter configurations) with other methods}
	\begin{tabular}[t]{|c|c|c|}
		\hline
		\textbf{Model} & \textbf{MSE-f} & \textbf{MSE-p}  \\ \hline
		EQ-PLLI (H=4, W=1) & 5.76& 15.10\\ \hline 
		OP-PLLI (H=4, W=1) & 3.40& 10.05 \\ \hline 
		PLLI (H=1, W=4) & 8.80& 11.32 \\ \hline 
		EQ-PLLI (H=2, W=2) & 5.83& 13.01\\ \hline 
		OP-PLLI (H=2, W=2) & 6.40& 12.40\\ \hline 
		Linear   model             & 16.79& 21.78 \\ \hline
		Regression tree explainer \cite{bastani2017interpreting} &21.17 & 37.27 \\ \hline
	\end{tabular}
	\label{tabl5:fid and mse boston}
\end{table}	
%


\subsubsection{Piecewise Local-Linear Interpreter}

There are several possible configurations to use for the piecewise interpreter. We  fix $K=4$ (as the dataset is small).  We have three parameter configurations possible ($H=4, W=1$), ($H=1, W=4$), ($H=2, W=2$). We compare the MSE-f and MSE-p in the Table \ref{tabl5:fid and mse boston}.  We use MSE-f to select among the various PLLI models with different configurations. We select OP-PLLI model with $(H=4, W=1)$.    In Table \ref{tabl5:fid and mse boston}, we compare PLLI with other models. We fit a regression tree (with four leaves as $K=4$) based explanation model  from \cite{bastani2017interpreting} and a linear regression model  to predict the black-box model and find that these approaches do not perform well. 


\subsubsection{Black-Box Model Interpretation}
We present the region-wise feature importance in Table \ref{tabl6: model summary boston}.  The table's different rows shows the different regions in the partition and the importance associated with different features. For instance, the fourth region has CRIM as the most important feature in contrast to the other regions where CRIM does not feature in the top five features.

\begin{figure*}[!h]
	\centering
	\begin{minipage}{.24\textwidth}
		\centering
		\includegraphics[width=2in]{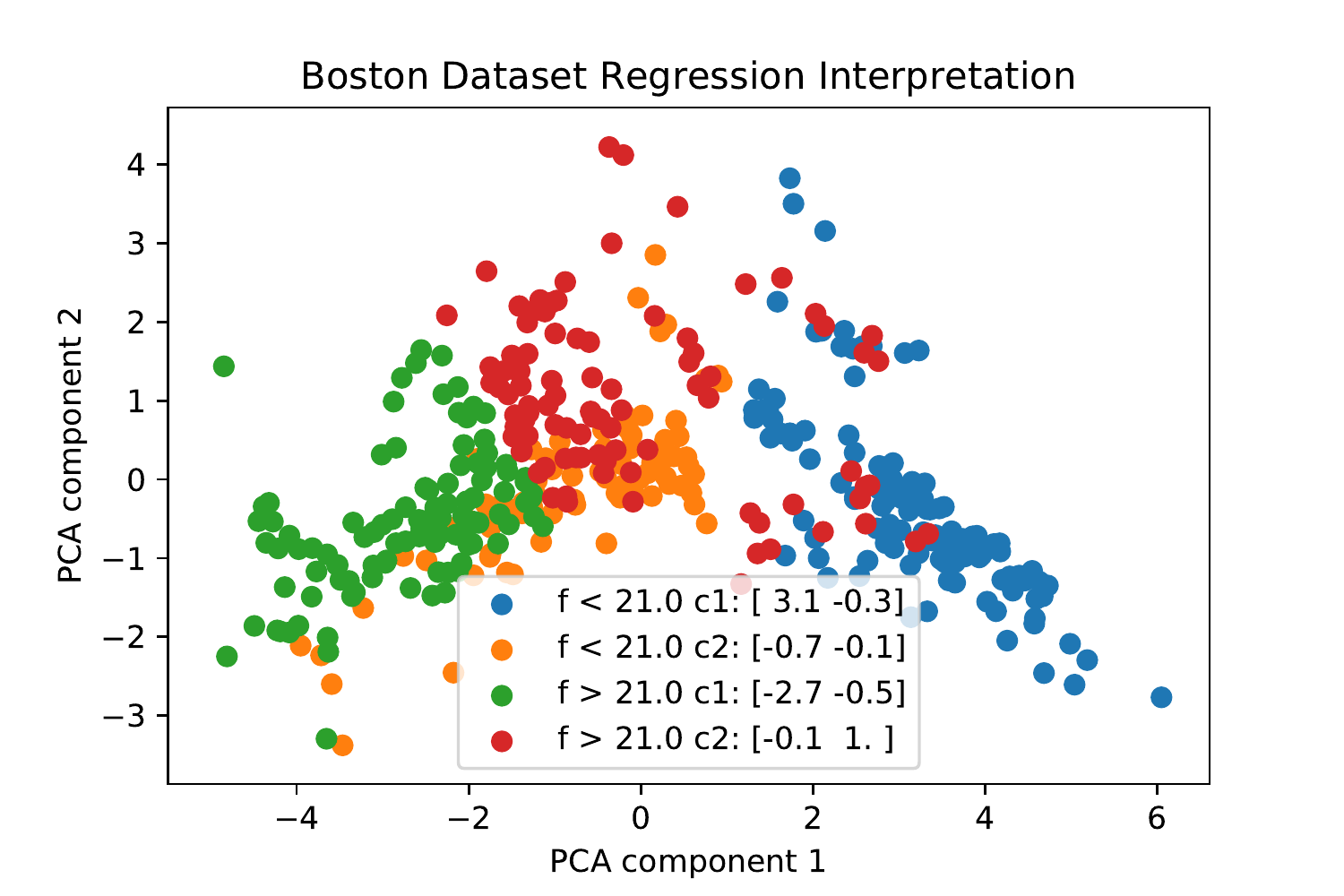}
		\subcaption{H=2, W=2, EQ}
		\label{fig:test1_b}
	\end{minipage}%
	\begin{minipage}{.24\textwidth}
		\centering
		\includegraphics[width=2in]{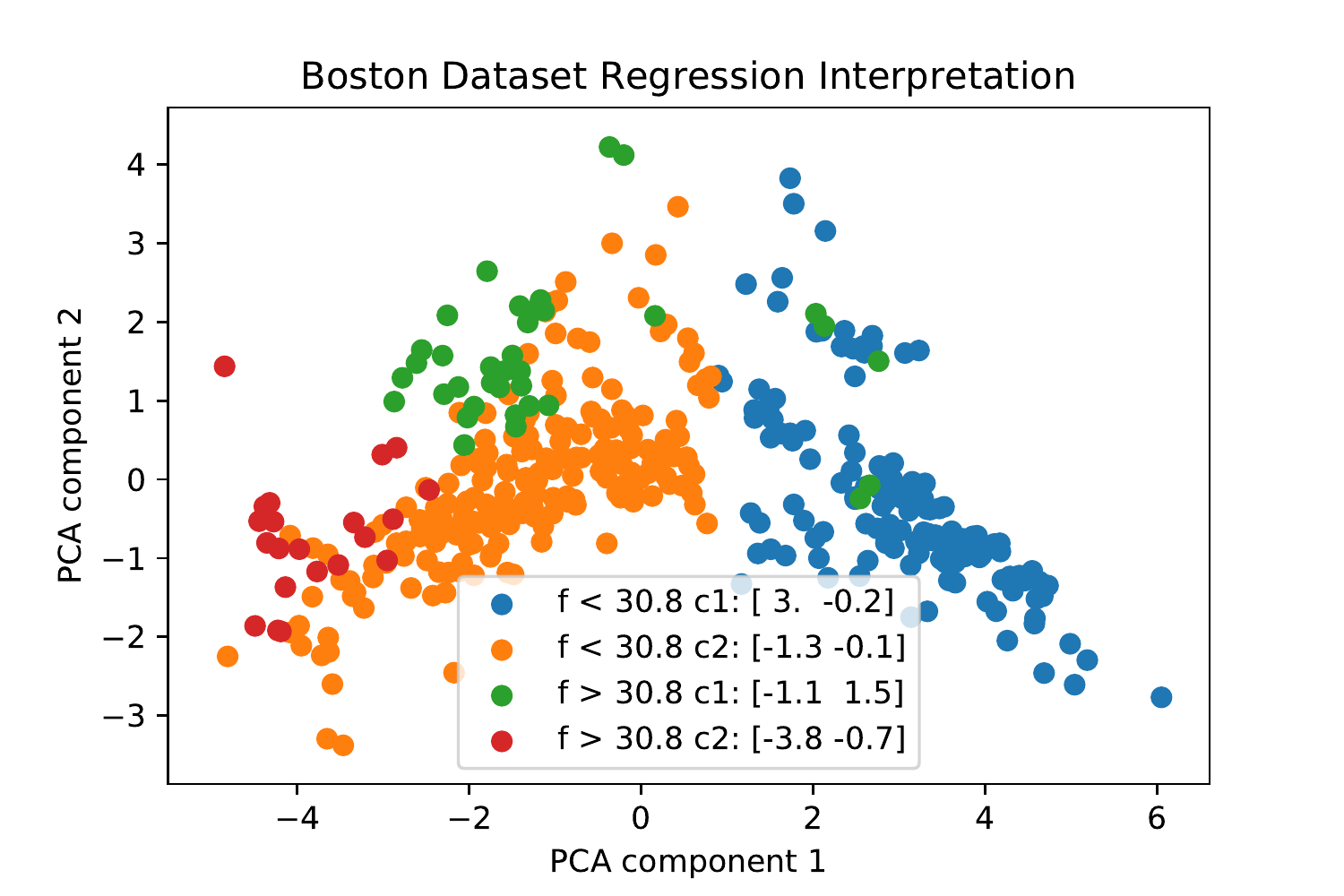}
		\subcaption{H=2, W=2, OP}
		\label{fig:test2_b}
	\end{minipage}
	\begin{minipage}{.24\textwidth}
		\centering
		\includegraphics[width=2in]{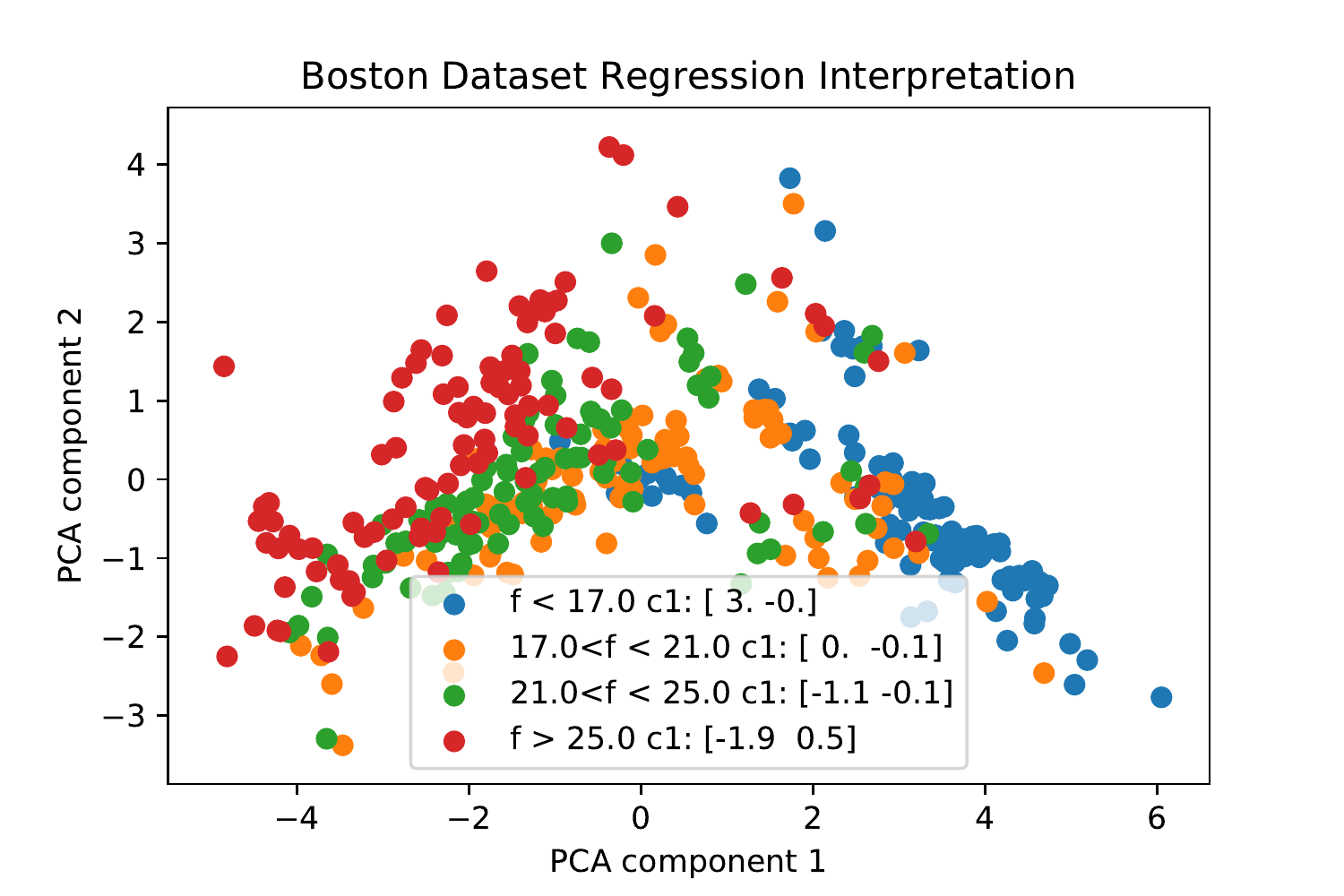}
		\subcaption{H=4, W=1, EQ}
		\label{fig:test3_b}
	\end{minipage}
	\begin{minipage}{.24\textwidth}
		\centering
		\includegraphics[width=2in]{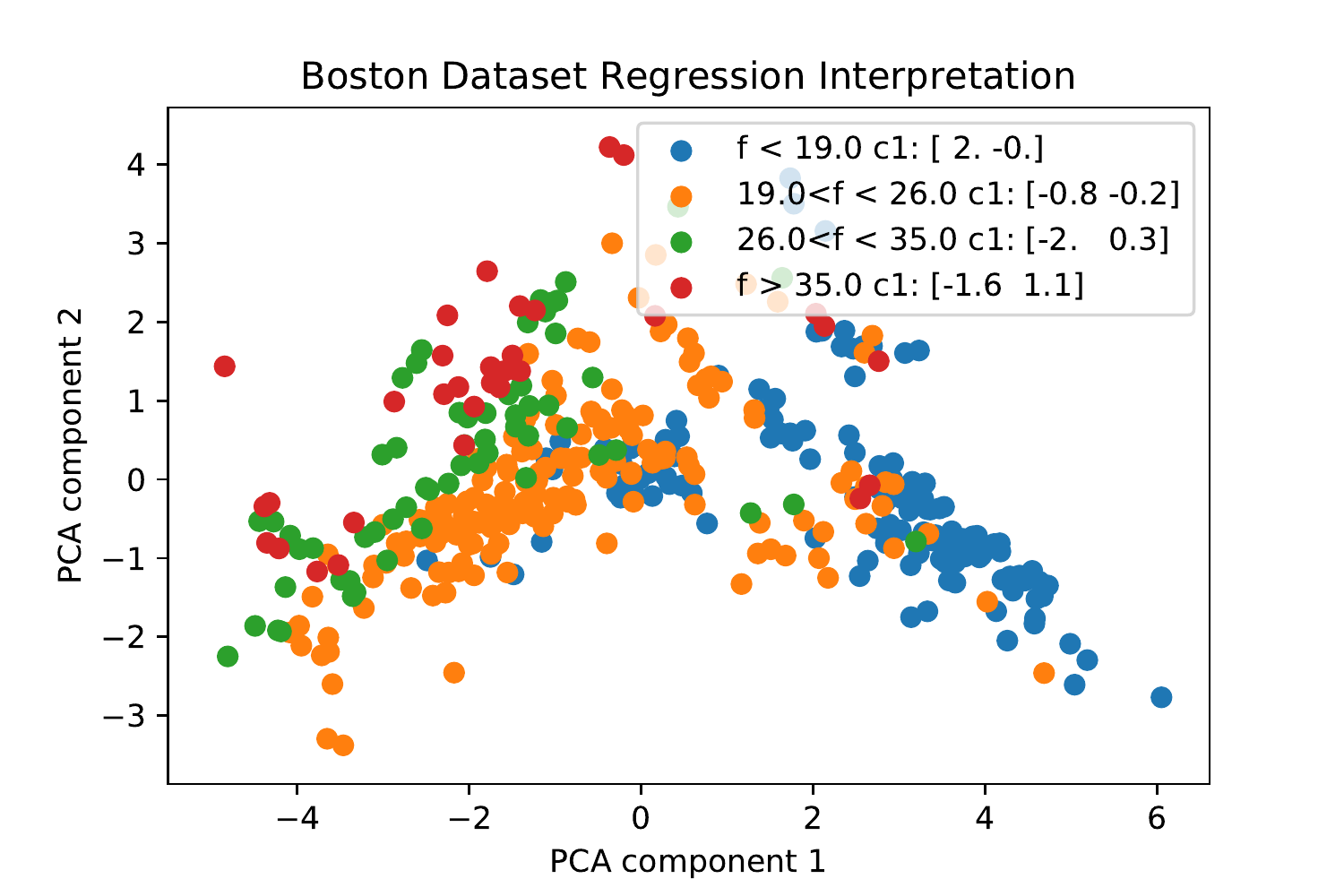}
		\subcaption{H=4, W=1, OP}
		\label{fig:test4_b}
	\end{minipage}
	\caption{Comparison of PLLI for different hyperparameter configurations}
	\label{fig6: comparison PI }
\end{figure*}

\begin{figure}[!htb]
	\begin{center}
		
		\includegraphics[trim= 0mm 0mm 0mm 0mm,  width=2 in]{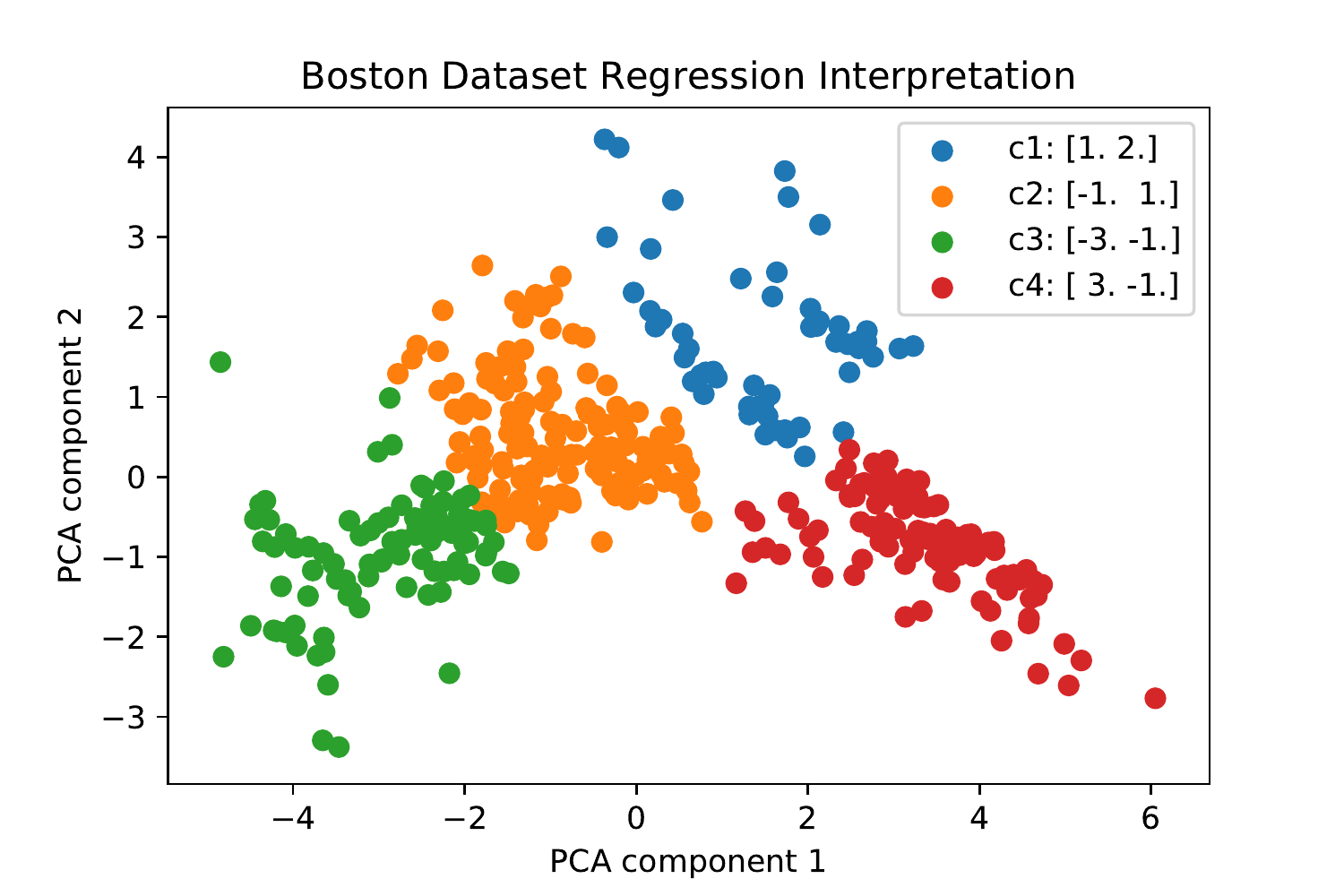}
		\caption{H=1, W=4, OP}
		\label{fig7: comparison PI}
	\end{center}
\end{figure}

 In Figure \ref{fig6: comparison PI } and \ref{fig7: comparison PI}, we show the different regions in the partitions. We compute PCA for the input features with two components. We use the two components of the PCA to represent the features. Based on the different parameters we get different partitions. In Figure \ref{fig6: comparison PI } and \ref{fig7: comparison PI}, the regions in the partitions are contiguous. The partition in Figure \ref{fig6: comparison PI } b) and Figure \ref{fig7: comparison PI} offers the additional advantage that the different data points in the partition do not overlap in the two dimensional space. If the data points do not overlap in the two dimensional space, then the partition is easy to visualize even in the 2-D plane.
 
\begin{table*}[!h]
	\renewcommand{\arraystretch}{1.1}
	\centering
	\caption{Region-wise feature importance. Top five features are highlighted in bold.}
	\begin{tabular}[t]{|c|c|c|c|c|c|c|c|c|c|c|c|c|c|}
		\hline
		\textbf{Region} & \textbf{CRIM} & \textbf{ZN} & \textbf{INDS} & \textbf{CHAS} & \textbf{NOX}   & \textbf{RM} & \textbf{AGE} & \textbf{DIS} & \textbf{RAD} & \textbf{TAX} & \textbf{PTR} & \textbf{B} & \textbf{LST}     \\ \hline
		$f<19.0, \; [2.0, 0.0]$ &0.63& \textbf{0.79} &0.68  &0.41&  \textbf{1.37}  & 0.14& 0.05&  0.31& 0.73 &  \textbf{1.46}&\textbf{1.04}  &0.18 & \textbf{1.59}\\ \hline		
		$19<f<26.0, \; [-0.8, -0.2]$ &0.40&0.17 &0.11  &0.22&   0.08 & \textbf{1.38}& 0.61& \textbf{0.63} & \textbf{0.94 }& \textbf{0.72}  &0.58  & 0.28& \textbf{1.43} \\ \hline		
		$26.0<f<35.0, \; [-2.0, 0.3]$ &0.0& 0.62& \textbf{1.39}  &0.10&0.16    & 0.95&0.57 &  0.94&\textbf{2.75}  & \textbf{1.57}  & \textbf{0.99} & 0.0&  \textbf{2.27} \\ \hline		
		$f>35.0, \; [-1.6, 1.1]$ &\textbf{6.61}
		& 0.03& 2.51 &0.08& \textbf{4.69}  &2.66 & 1.34& 2.71 &\textbf{7.74}  &\textbf{3.04}  & \textbf{3.02}  & 0.0& 2.71 
		\\ \hline		
	\end{tabular}
	\label{tabl6: model summary boston}
\end{table*}	

\subsubsection{Identifying data points for local explanations}
In \cite{ribeiro2016should}, the authors proposed a method to identify candidate data points at which they give local instance based explanations in order to provide a somewhat global view of the model behavior. The  method in \cite{ribeiro2016should} is called Submodular Pick. It tries to ensure that data points that are selected present a diverse set of feature importance. However, the selected data points do not necessarily represent a diversity in terms of the feature distribution or the predicted-value distribution. A naive approach to select the data points is to select them randomly. 

Our method provides a natural way to select the candidate data points. The data points selected by our method are the centroids of each region in the partition identified by the PLLI.
Suppose we want to identify $K$ candidate data points, in that case, we set the size of the partition for PLLI as $K$. At each of these data points we use the  method in \cite{ribeiro2016should} to construct instancewise explanations. We can also use other local explanation frameworks (See Related Works Section) to provide these instancewise explanations. Hence, we see that our framework nicely complements the existing local explanation frameworks.

\begin{figure*}[!h]
	\begin{minipage}{.48\textwidth}
		\centering
		\includegraphics[width=3.8in]{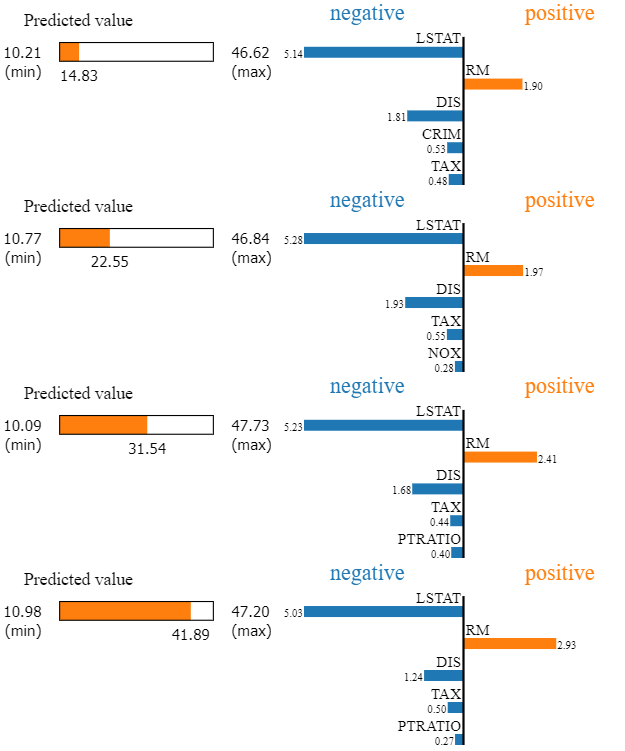}
		\label{fig:ex_p}
		\subcaption{}
	\end{minipage}
	\begin{minipage}{.48\textwidth}
		\centering
		\includegraphics[width=1.8in]{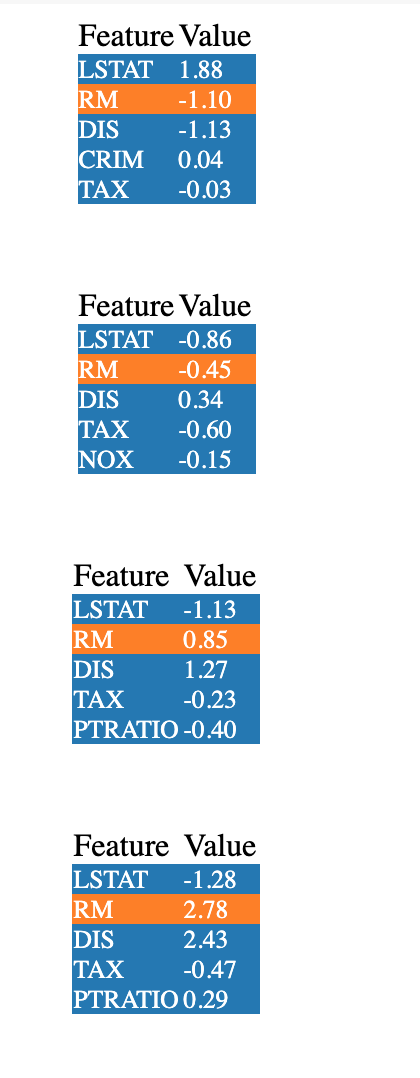}
		\label{fig:ex_f}
		\subcaption{}
	\end{minipage}
	\caption{Data points selected based on PLLI and corresponding explanations (for top five features)}
	\label{fig8: example}
\end{figure*}

 Next, we want to compare of our approach against submodular pick and random selection. First, we define how well spread are the data points as follows. For each selected point compute the distance from the nearest neighbor. We define coverage as the average minimum distance from the neighbors, i.e., $\frac{1}{K}\sum_{i}\min_{j \not = i} \|x_i-x_j\|$). We measure the coverage for the feature importances, the coverage for the feature vectors, and the coverage for function values. We compare the proposed procedure (for $K=4$) with random method (averaged over 10 runs) and the submodular pick method. In Table \ref{table7: coverage}, we compare the various methods. We observe that the proposed method is better at giving a larger coverage in terms of feature values, predictions and importance values. In Figure \ref{fig8: example}, we show the different representative data points (predicted values, explanations and features) identified by the proposed method. 



\begin{table}[!h]
	\renewcommand{\arraystretch}{1.2}
	\centering
	\caption{Comparison of coverage of various selection methods}
	\begin{tabular}[t]{|c|c|c|c|}
		\hline
		\textbf{Algorithm} & \textbf{Coverage } & \textbf{Coverage}
		 & \textbf{Coverage }  \\ 
		 	 & \textbf{ importances} & \textbf{ predictions} 	 & \textbf{features}  \\ \hline
		PI & 0.68& 8.69&3.80\\ \hline 
		Submodular pick & 0.63& 4.38 &3.20\\ \hline 
		Random& 0.60& 3.87 & 3.75\\ \hline 
		
	\end{tabular}
	\label{table7: coverage}
\end{table}	

\subsection{Large dataset}
The computational complexity of the proposed approach is large (See Proposition 3). It is easy to approximate the PLLI procedure (described in the Appendix) to allow it to scale to large datasets.
In this section, we show that an approximate version of PLLI can scale well for large datasets as the  experiments in the previous sections were done on datasets that were moderately small (500-1000 datapoints). In this section, we  use the California Housing Dataset from StatLib library below. It consists of 20,640 data points with 8 features. The 8 features are described as 
\begin{itemize}
	\item MedInc: median income in block
	\item HouseAge: median house age in block
	\item AveRooms: average number of rooms
	\item AveBedrms: average number of bedrooms
	\item Population: block population
	\item AveOccup: average house occupancy
	\item Latitude: house block latitude
	\item Longitude: house block longitude. 
\end{itemize}

The target variable is the median house value for California districts. We fit a RF regressor to predict the target variable. In Table \ref{table8}, we compare the RF regressor with more interpretable methods.
We compare the performance of EQ-PLLI $(H=4, W=1)$ with approximate OP-PLLI $(H=4, W=1)$. The goal is to show that with a reasonable computation time the proposed approximation approach performs well.   
In Table \ref{tabl9:fid and mse california}, we compare approximate OP-PLLI method with EQ-PLLI. Note that approximate OP-PLLI took 700 seconds to train, while the exact OP-PLLI would take 3 days to train. 

\begin{table}[!h]
	\renewcommand{\arraystretch}{1.2}
	\centering
	\caption{Comparison of RF Regressor with other methods}
	\begin{tabular}[t]{|c|c|c|}
		\hline
		\textbf{Model} & \textbf{MSE-p} & \textbf{R}$^2$  \\ \hline
		RF Regressor & 0.24& 0.80 \\ \hline 
		Regression tree & 0.74& 0.58 \\ \hline 
		Linear model & 0.53& 0.60 \\ \hline 
	\end{tabular}
	\label{table8}
\end{table}	

\begin{table}[!h]
	\renewcommand{\arraystretch}{1.2}
	\centering
	\caption{Comparison of PLLI interpreter different hyperparameter configurations}
	\begin{tabular}[t]{|c|c|c|}
		\hline
		\textbf{Model} & \textbf{MSE-f} & \textbf{MSE-p}  \\ \hline
		EQ-PLLI (H=4, W=1) & 0.084& 0.33\\ \hline 
		Approximate-OP-PLLI (H=4, W=1) & 0.076& 0.26 \\ \hline 
	\end{tabular}
	\label{tabl9:fid and mse california}
\end{table}

\section{Connection with k-means clustering}
In this section, we begin by drawing a connection between the equation \eqref{minim-prob} and the general problem of clustering.  


\subsection{Ordered Partitions} 

We say that the partition $\mZ$ of $A \subset \mX$ is {\em ordered} if for every $Z, Z' \in \mZ$ with $Z\not=Z'$, either

(i) for all $z \in Z, z' \in Z'$ we have $f(z) < f(z')$, or 

(ii) for all $z \in Z, z' \in Z'$ we have $f(z) >f(z')$

In this section, we consider the same setting as in the Section \ref{Main_results}, where $H=K$ and $W=1$, i.e., we only want to optimize how to divide the function's range.  Hence, we only search in the space of $\{a_r\}_{r=1}^{H-1}$, which characterizes the different intervals that are possible. Recall that we define the set of partitions that we search in as $\mathcal{P}_{K}(\mathcal{X})^{\dagger}$. Consider any two regions $Z_i=f^{-1}[a_m, a_n]$ and $Z_j =f^{-1}[a_p, a_q]$ in partition $\mathcal{Z} \in \mathcal{P}_{K}(\mathcal{X})^{\dagger}$. $[a_m, a_n]$ and $[a_p, a_q]$ are non-overlapping intervals by construction. Hence, the two regions $Z_i$ and $Z_j$ are ordered. Therefore, $\mathcal{P}_{K}(\mathcal{X})^{\dagger}$ is the set of all the ordered partitions of size  $K$. 

Next  we reformulate the problem \eqref{minim-prob}. We search in the space of locally constant models instead of linear models \eqref{minim-prob}.  We expand the search to the space of all the partitions of size  $K$, $\mathcal{P}_{K}(\mathcal{X})$, instead of ordered partitions in \eqref{minim-prob}. 
We formally state the assumption on local models as 

\textit{Assumption 4:} The family of local models $\mathcal{H}$ are constant models.

 The reformulated problem is  stated as 

\begin{equation}
(\mathcal{M}^{\uparrow}, \mZ^{\uparrow}) = \argmin_{\mathcal{M} \in \mathcal{H}^{K},  \mZ \in \mP_K(\mX)}  \hat{R}(\mathcal{M}, \mathcal{Z}; D) 
\label{minim-prob1}
\end{equation}

We simplify the above problem in \eqref{minim-prob1}. Suppose $\{Z_1,..,Z_K\}$ are the regions in the partition and $\{c_1,...,c_k\}$ are the corresponding locally constant model values. We write the loss as follows 
\begin{equation}
\hat{R}(\mathcal{M}, \mathcal{Z}; D)  = \frac{1}{n} \sum_{k=1}^{K}\sum_{x_i \in Z_k} \big[l(|f(x_i) - c_k|) \bigr]   
\label{emp-risk1}
\end{equation}
Let $f(x_i)= y_i$. Suppose $\mathcal{I}= \{I_1,...,I_K\}$ is a partition of $\{y_1,...,y_n\}$ in $K$ regions. Observe that $\mathcal{I} \in \mathcal{P}_K(\mathcal{Y})$, where $\mathcal{P}_K(\mathcal{Y})$ is the partition of $\mathcal{Y}$ in $K$ regions.  
We  rewrite \eqref{emp-risk1} as
\begin{equation}
\hat{R}(\mathcal{M}, \mathcal{I}; D)  = \frac{1}{n} \sum_{k=1}^{K}\sum_{y_i \in I_k} \big[l(|y_i - c_k|) \bigr]   
\label{emp-risk2}
\end{equation}

We formulate risk minimization with  \eqref{emp-risk2} as objective

\begin{equation}
(\mathcal{M}^{\downarrow}, \mathcal{I}^{\downarrow}) = \argmin_{\mathcal{M} \in \mathcal{H}^{K},  \mathcal{ I} \in \mP_K(\mY)}  \hat{R}(\mathcal{M}, \mathcal{I}; D) 
\label{minim-prob2}
\end{equation}
Observe that the two optimization problems \eqref{minim-prob1} and \eqref{minim-prob2} are equivalent. It is easy to show this using  contradiction.  Also, observe that the optimization problem \eqref{minim-prob2} is a general clustering problem for one-dimensional data $\{y_1,..., y_n\}$.   Recall that if we set $\ell$ to be a squared function in \eqref{minim-prob2}, we obtain MSE minimization. The classic k-means clustering method \cite{jain2010data} also tries to solve this problem. 

Next, we discuss how to solve the problem in \eqref{minim-prob1}. We use $\{y_1,..., y_n\}$ as input and constant local models in the Algorithms 1 and 2. We set $H=K$ and $W=1$ in Algorithm 1 and 2. Next, we  prove that the output of Algorithm 1 and 2 achieves the optimal clustering in polynomial time.  We first begin by getting worst case complexity bounds for the Algorithm 1 and 2.
\begin{proposition} \textit{Computational Complexity:}
	If Assumption 1,3, and 4 hold, then 	the computational complexity of Algorithm 1 and 2 together is $\mathcal{O}(|D|^3 K)$.
\end{proposition} 

We give the proof to this Proposition in the Appendix.


In the Proposition 3, we proved the complexity for mean squared error loss function.  What can we say about the complexity for a general $\ell$?. If $\ell$ is a square function and the local model is a constant model, then computing  $G(i,j)$ in \eqref{eqn: centroid} has the worst case complexity of  $\mathcal{O}(|j-i|)$.  If $\ell$ is any general  convex function, then it is hard to have analytical expression for the worst case complexity of computing $G(i,j)$. \footnote{It is possible to arrive at a bound on number of steps if $\ell$ is self-concordant, see \cite{boyd2004convex} for details} However, since $\ell$ is convex computing $G(i,j)$ amounts to solving a convex minimization problems.  Hence, Algorithm 1 and 2's complexity amounts to solving $|D|^2K$ convex optimization problems. 

In the next proposition, we state that the output of Algorithm 1 and 2 achieves optimal clustering.

\begin{proposition} \textit{Optimal Clustering:}
If Assumption 1,3, and 4 hold, then the output of the Algorithm 1 and 2 achieves optimal clustering, i.e.,
	$  \hat{R}(\mathcal{M}^{\#},\mZ^{\#}; D) = \min_{\mathcal{M} \in \mathcal{H}^{K},  \mathcal{ I} \in \mP_K(\mY)}  \hat{R}(\mathcal{M}, \mathcal{I}; D)  
	$	
\end{proposition} 

The Proof of Proposition 4 is in the Appendix Section. Note that we have established that the Algorithm 1 and 2 achieves optimal clustering in polynomial time. Methods in the literature such as k-means clustering \cite{jain2010data} are not guaranteed to achieve the optimal clustering even for the one-dimensional case that we described above. When $\ell$ is a general loss function, then the problem in \eqref{minim-prob2} is the general clustering problem. In the Appendix Section, we state the generalization of the above proposition, which states the outcome of Algorithm 1 and 2 solve the general clustering problem. The proof of the generalization  uses ideas from the theory of influence functions \cite{cook1982residuals}.

\section{Related Work}



The related works on model interpretation and piecewise model approximation can be broadly categorized into two categories. The first kind provides a more global explanation (our work falls in this category); the second kind provides a local explanation for instance, explain the predictions made for a certain data point. We give the most representative works in each category but by no means this is exhaustive.

\subsection{Global frameworks:}
%

\subsubsection{Tree based approximations } In \cite{bastani2017interpreting},  the authors develop an approach to approximate the entire black-box function.  The objective of their work is to find the best tree that approximates the black-box model. In the authors approach, they sample data points based on active sampling from estimated feature distribution and then constructed a decision tree to approximate the black-box model.  This tree construction is based on a greedy algorithm, which can lead to poor approximation (In our experiments, we showed that this approach leads to poor approximations and hence unreliable interpretations.)

\subsubsection{Decision set based approximation } Decision sets (sets of if-then rules)  based black-box approximations have been used in \cite{lakkaraju2017interpretable}. 
The framework in \cite{lakkaraju2017interpretable} optimizes an objective that balances the ambiguity between the classifier and the decision rule against the interpretability of the decision set. The approach of  \cite{lakkaraju2017interpretable}  only constructs piecewise constant approximations and does not provide learning guarantees. 

\subsubsection{Piecewise Interpolation} In this  section, we would also like to make an important connection between this work and other piecewise interpolation methods such as radial basis function networks \cite{orr1999recent}, spline-based interpolation \cite{friedman1991multivariate}. In radial basis function networks, the output of the network can be considered as an average over different models defined at each of the centers. In these methods, the centers are usually selected randomly or in an unsupervised manner, which is why the regions constructed by these methods may not be homogeneous, unlike our approach. In contrast in our approach,  these centers are learned. MARS \cite{friedman1991multivariate} is an adaptive splines method that constructs piecewise models. MARS uses a recursive partitioning based approach to partition the feature space. Recursive partitioning uses a greedy procedure, which can often lead to poor approximations especially when the tree is not deep.


\subsection{Local frameworks}
While our method falls in the category of global framework, we give a description of local frameworks for completion.   These frameworks \cite{ribeiro2016should} \cite{shrikumar2017learning} \cite{lundberg2017unified} \cite{koh2017understanding} produce a linear approximation of the black-box model in some  neighborhood of a given point.  The coefficients of the linear approximation represent the importance associated with different features for a certain data instance. We provide more specific details for these different frameworks below.  Instancewise feature importance methods \cite{chen2018learning} fall in the category of local frameworks. We already covered their description in the Introduction Section.

\subsubsection{LIME}  In the LIME framework \cite{ribeiro2016should},  the linear approximation of the black-box at each input data point is computed using weighted least squares, with weights coming from an exponential kernel.  The authors also extend the framework to identify a set of points at which to provide  local-linear approximations; this helps with global model explanation. The points that are identified are not guaranteed to be diverse enough (further explained in the Experiments Section), i.e., they may all come from similar regions in the feature space and/or the range space of the black-box function.


\subsubsection{Relevance propagation} Layerwise relevance propagation (LRP) \cite{bach2015pixel} and  DeepLIFT \cite{shrikumar2017learning}  are relevance propagation frameworks commonly used for interpretation of neural networks. In DeepLIFT, the importance of a feature is determined by perturbing the feature of a data point from a reference value and tracking the gradient/change in the outcome/prediction. This work is particularly designed to handle the problems that arise in neural networks such as gradient saturation that are not handled by others \cite{selvaraju2016grad}.

\subsubsection{SHAP} The SHAP framework \cite{lundberg2017unified}  shows that many of the existing frameworks can be understood as variants of a common linear interpretive model. The authors argue that the coefficients of a good linear interpretive model should satisfy a particular set of axioms and shows that the coefficients derived from Shapley regression provide the unique solution to these axioms.  


%

\subsection{Connection between piecewise local-linear models and existing local frameworks}
In this work, we have extensively looked at piecewise linear models. Existing local frameworks described in the previous section construct a local model for each data instance. In most cases, these local models are linear \cite{lundberg2017unified}. Since there is one model per data instance these models qualify as piecewise local-linear models, where every different model corresponds to a different piece in a piecewise model.
For a new data instance, we find the nearest data point in the training set and use the corresponding local-linear model to find the interpretation for the new data instance.
In general,  it is impractical to have a different model to explain every instance and inspect each of those interpretations. It is more practical to have a model that represents a group of data points keeping the total number of pieces required to explain the model below a reasonable value, which is what our method achieves.



\section{Conclusion} 
This paper provides a novel  way to construct piecewise approximations of a black-box model. Our approach uses dynamic programming to  partition the feature space into regions and then assigns a simple local model within each region.  We carry out experiments to show that the proposed approach achieves a smaller loss and better reflects the black-box model compared to other approaches. We also show how the proposed approach can be used for a) computing region-wise feature importance, and b) identifying representive data points for instancewise explanations.  We also prove that the proposed approach can also be applied to the problem of clustering. We provide a first proof that the proposed approach achieves optimal clustering in polynomial time when the data is one-dimensional.
\section*{Acknowledgement}
We would like to acknowledge Professor Gregory Pottie (Department of Electrical and Computer Engineering, UCLA) for valuable comments that helped us improve the paper. We would like to acknowledge the  Office of Naval Research (ONR) and the National Science Foundation (NSF) Award 1524417 for supporting this work. Kartik Ahuja would like to acknowledge the support from the Guru Krupa Fellowship Foundation.

\appendix


\section*{Proof of Proposition 1}

\textbf{Bellman Principle} Let $\mZ$ be an ordered partition of $\mX$ and assume that $R(f, g_{\mathcal{M}, \mathcal{Z}}, D)$ minimizes the risk among all ordered
partitions of 
$\mX$ with at most $|\mZ|$ elements.  If $\mZ', \mZ'' \subset \mZ$ is a partition of $\mZ$, then $R(f, g_{\mathcal{M}, \mathcal{Z}^{'}}, D)$ minimizes the risk among all ordered partitions of $A'$ with at most $|\mZ'|$ elements.  (If this were not true then we could find another ordered partition $\mZ^*$ of $A'$ with lower risk.  But then $\mZ^* \cup \mZ''$ would be an ordered partition of $\mX$ with lower risk than $\mZ$, which would be a contradiction.) We now use this principle to express that the optimal risk satisfies the  Bellman equation. Suppose that the first $n$ points are to be divided into $k$ intervals.  The minimum risk achieved by the optimal partition of first $n$ points into $k$ intervals $V(n,k)$ satisfies the Bellman equation given as. 

$$V(n+1,k) = \min_{n^{'} \in \{1,..,n\}}\{V(n^{'},k-1)+ G(n^{'}+1,n) \}$$

Note that $V(1,k)=0$ for all $1 \leq k\leq K$ (the model $M$'s output is equal to the label of one data point itself). We refer to  $V$ as value function.

To prove this proposition, we first state a lemma.
\begin{lemma}
	The value function output by the Algorithm 1 is the same as the true value function $V$, i.e., $V^{'} = V$. 
	
\end{lemma}

\subsection*{Proof of Lemma 1.}


We use induction in the number of data points $n$. 

We start with the base case $n=1$. For $n=1$ and $1 \leq k \leq K$, we know that the $V^{'}(1,k)=0$ (from the initialization of the Algorithm).  We also know that $V(1,k)=0$ for all $1 \leq k\leq K$ (the model $M$'s output is equal to the one data point itself).   Hence, the claim is true for $n=1$ and for all $1 \leq k\leq K$.

Suppose that the Algorithm outputs optimal value functions for all $s\leq n$ and for all $k\leq K$. 

Consider the data point $n+1$ and the constraint on the number of partitions is $k$. From the Algorithm 1 we know that $$V^{'}(n+1,k) = \min_{n^{'} \in \{1,..,n\}}\{V^{'}(n^{'},k-1)+ G(n^{'}+1,n) \}$$

Let us assume that $V^{'}(n+1,k)$ is not optimal, i.e., $$V^{'}(n+1,k) >V(n+1,k)$$ We use the Bellman principle to write the value function $V$ as follows
$$V(n+1,k) = V(n^{*},k-1) + G(n^{*}+1,n)$$
We use the above two equations to write 
$$V^{'}(n+1,k) > V(n^{*},k-1) + G(n^{*}+1,n)   $$ 

We also know that 
$$V^{'}(n+1,k) < V^{'}(n^{*},k-1) + G(n^{*}+1,n)   $$ 

Therefore, we can write 

$$V^{'}(n^{*},k-1) + G(n^{*}+1,n) > V(n^{*},k-1) + G(n^{*}+1,n)$$
$$ \implies V^{'}(n^{*},k-1) > V(n^{*},k-1) $$


This contradicts the assumption that $V(n^{*},k-1)  =V^{'}(n^{*},k-1) $. Hence, the assumption $V^{'}(n+1,k)> V(n+1,k)$ cannot be true, which implies $V^{'}(n+1,k)\leq V(n+1,k)$. Thus we can say that $V^{'}(n+1,k)= V(n+1,k)$ ($V^{'}(n+1,k)<V(n+1,k)$ can't be true since $V$ is optimal value function).  \hfill QED

In Lemma 1, we showed that $V^{'}=V$.   To complete the proof of Proposition 1, we need to show that the partition output by the Algorithm 2 achieves $V$.

Recall the computation of value function from Algorithm 1 $$V^{'}(n+1,k) = \min_{n^{'} \in \{1,..,n\}}\{V^{'}(n^{'},k-1)+ G(n^{'}+1,n) \}$$

From Algorithm 1, we also know that 
$$\Phi(n+1,k) = \argmin_{n^{'} \in \{1,..,n\}}\{V^{'}(n^{'},k-1)+ G(n^{'}+1,n) \}$$
We can write $$V^{'}(n+1,k) = V^{'}(\Phi(n,k),k-1) + G(\Phi(n,k)+1,n)$$ 

The subset of the data up to data point $n$ is written as $D_n$. The optimal partition with $n+1$ points and at most $k$ regions induces a partition of the dataset $D_{n+1}$. We write the last region of the induced partition on $D_{n+1}$ as $S_{k}$. We know that  $S_{k} = \{\Phi(n,k)+1,..,n\}$. 

We can repeat this procedure recursively and define $S_{k-1}$ and so on. The set of points $ S_{k-1} = \{\Phi(\Phi(n,k), k-1),..., \Phi(n,k)\}$ is the set of points that belong to the region $k-1$ and so on. This induced partition achieves the risk value of $V^{'}(n+1,k)$

We require that the partition constructed in Algorithm 2 also divides the points in the dataset in the exact same manner as described above. 
%
%

Consider the region $Z_K$ output by the Algorithm 2. 
$$Z_K = \{x:f(x_{\Phi(|D|,K)}) <  f(x) \leq f(x_{|D|}) \}$$

The points $\{\Phi(|D|,K)+1,..,|D|\}$ are ordered and thus all of these belong to $Z_K$.  The same argument applies to the set $Z_{K-1}$ (given below)   and the set of points  $\{\Phi(\Phi(|D|,K),K-1)+1,..,\Phi(|D|,K)\}$ and so on. 
\begin{equation}
\begin{split}
Z_{K-1} = \{x:||f(x_{\Phi(\Phi(|D|,K),K-1)})|| <  ||f(x)|| \leq ||f(x_{\Phi(|D|,K)})|| \}
\end{split}
\end{equation}

Observe that the division of the points is the same as prescribed by the value function $V^{'}$. Hence, the output of Algorithm 2 achieves $V^{'}$ and from Lemma 1 we know it is equal to the minimum risk.

\section*{Proof of Proposition 2}

From Proposition 1,  we know that Algorithm 1 and 2 combined solve the empirical risk minimization (ERM) problem. In this proposition, we claim that the ERM actually leads to succesful agnostic PAC learning. 

Consider the MSE loss. The feature space is $\mathcal{X}$ and the label space is $\mathcal{Y}$. We consider a discretization of the label space $\mathcal{Y}_d$. This  discretization trick is fairly common see \cite{shalev2014understanding}. 

If $\mathcal{Y}=[0,1]$, then a quantization of $\mathcal{Y}$ into steps of length $\Delta$ is given as $\mathcal{Y}_d = \{0,\Delta,2\Delta....1\}$. 
Suppose the partition of the dataset is $\{D_1,..., D_K\}$ and the corresponding set of local-linear models are $\{M_1,..., M_K\}$. Each $M_k: \mathcal{X} \rightarrow \mathcal{Y}$ (If we use a local-linear model we can clip the linear function and keep it bounded between $[0,1]$).
We write the MSE loss for the continuous labels/black-box predicted values as $$MSE_c = \frac{1}{|D|}\sum_{k=1}^{K} \sum_{x \in D_k}(y-M_k(x))^2$$
where $y=f(x)$.
Suppose the discretization of a label $M_k(x)$ is $y_d \in \mathcal{Y}_d$ (we discretize a point by finding the closest point from $\mathcal{Y}_d$). 
We write the discretized MSE as 
$$MSE_d = \frac{1}{|D|}\sum_{k=1}^{K} \sum_{x \in D_k}(y-y_d)^2$$
We compare these MSEs 
\begin{equation}
\begin{split}
|MSE_c - MSE_d |=  \frac{1}{|D|}\sum_{k=1}^{K} \sum_{x \in D_k}|(2y-y_d-M_k(x))||M_k(x)-y_d | \leq 4\Delta 
\end{split}
\label{mse_diff_o}
\end{equation}

The same applies to the expected MSE as well.
\begin{equation}
\begin{split}
&|E[MSE_c] - E[MSE_d ]|\leq 4\Delta 
\end{split}
\label{emse_diff_o}
\end{equation}

%

For the sake of clearer exposition, let us assume that the hypothesis class that we want to optimize over is of constant models $M_k(x)=c$ instead of linear models in \eqref{eqn:truerisk}. We will show at the end how to extend the proof to linear class.
The MSE loss $MSE_c$ is minimized when $\bar{y}_k =\frac{1}{|D_k|} \sum_{x \in D_k} f(x)$ and we write the loss w.r.t. $\bar{y}_k$. 
$$MSE_c^{*} =\frac{1}{|D|}\sum_{k=1}^{K} \sum_{x \in D_k}(y-\bar{y}_k)^2 $$
  


Suppose we  restrict the search of the minimizer $\bar{y}_k$ to discretized values as well. In that case, we take the discretization of the sample mean $\bar{y}_k$, which we denote as $\bar{y}_k^d \in \mathcal{Y}_d$  as the discretized optimal solution. We write the mean square error with $\bar{y}_k^d$ as

  $$MSE_c^{\dagger} = \frac{1}{|D|}\sum_{k=1}^{K} \sum_{x \in D_k}(y-\bar{y}_k^d)^2$$ 

The difference between the MSEs is given as follows. 

\begin{equation}
\begin{split}
&|MSE_c^{\dagger} - MSE_c^* |=  \frac{1}{|D|}\sum_{k=1}^{K} \sum_{x \in D_k}|(2y-\bar{y}_k-\bar{y}^{d}_k)|| \bar{y}_k^d-\bar{y}_k| \leq 4\Delta 
\end{split}
\label{mse_diff}
\end{equation}

If the discretization level $\Delta$ is sufficiently small,  finding the minimizer in the discrete space is almost as good as the minimization in the continuous space. 

We also want to discretize the partitions.  Next, we consider the effect of discretization.

In the above expression \eqref{mse_diff} each set $D_k$ is obtained based on the partition of the continuous label space $\mathcal{Y}=[0,1]$.  If we limit the search to the partitions based on the discreteized label space $\mathcal{Y}_d$, then also it is easy to show that the difference due to the quantization can be made arbitrarily small.  
Let us consider a region $D_k $ defined as follows. If $x\in D$ and  $f(x) \in [a,b] $, then $x \in D_k$. Let us consider the discretized version of $D_k$ defined as $D_k^d$. If $f(x) \in [a_d, b_d]$, then $x \in D_k^d$, where $a_d,b_d$ are discretizations of $a,b$.  Define the difference the total errors for the two regions as follows.
\begin{equation}
\begin{split}
&SE(D_k) - SE(D_k^d) =  \sum_{x \in D_k} (f(x) -\mu)^2  - \sum_{x\in D_k^d}(f(x)-\mu)^2 = \\ & \sum_{x\in D_k \cap [D_k^d]^c} (f(x)-\mu)^2 - \sum_{x\in D_k^c \cap D_k^d}(f(x)-\mu)^2
\end{split}
 \end{equation}
 where $A^c$ is the complement of set $A$.
 The difference between the loss defined over these partitions will vary the most when these partitions are maximally different, which happens when $a = a + \Delta$ and $b = b-\Delta$. 
 $$SE(D_k) - SE(D_k^d)  = \sum_{f(x)\in [a, a+\Delta] \cup [b-\Delta, b]} (f(x)-\mu)^2 $$
 Since $f(x)\in [0,1], \mu \in [0,1]$ $\implies $ $(f(x)-\mu)^2 \leq 1$. Therefore, $SE(D_k) - SE(D_k^d)$ is bounded above by the number of points in $[a, a+\Delta] \cup [b-\Delta, b]$. The total number of points in $[a, a+\Delta] \cup [b-\Delta, b]$ as the data becomes large is $\approx Pr(f(x) \in [a, a+\Delta] \cup [b-\Delta, b]) |D|$.  If $\Delta$ is very small and if we assume that the cumulative distribution function (c.d.f) of $f$ is continuous, then it follows that  $Pr(f(x) \in [a, a+\Delta] \cup [b-\Delta, b]) |D| = v(a)\Delta|D| + v(b)\Delta |D|$, where $v$ is the probability density function (p.d.f) of $f$.  Therefore, 
 \begin{equation}
 SE(D_k) - SE(D_k^d)  \leq (v(a) + v(b))\Delta |D|
 \label{SE_diff}
 \end{equation}
 From the above \eqref{SE_diff} it follows that the difference in the MSEs can be made arbitrarily small.
  Observe that the objective function in \eqref{eqn:truerisk} only consists of the values of $f$.  Also, observe that the partition of $\mathcal{X}$ creates a partition of the domain of $f$, i.e., $\mathcal{Y}$.  To solve \eqref{eqn:truerisk} one can equivalently search over all the ways to partition $\mathcal{Y}$ into $K$ intervals. Hence \eqref{eqn:truerisk} can be restated as 
  \begin{equation}
  \min_{\mathcal{M}\in \mathcal{H}^{K}, \mZ \in \mP_K(\mY)^{\dagger}} R(f, g_{M, \mZ}; \mD)  \label{eqn:truerisk_d1} 
  \end{equation}
  where $\mP_K(\mY)^{\dagger}$ is the set of all the partitions of $\mY$ into $K$ regions where each region is an interval.
  If we restrict the boundary of each interval to be from the set $\mathcal{Y}_d$, then the set of all the discretized partitions  $\mathcal{P}_{K}^d(\mathcal{Y})^{\dagger}$.

We reformulate \eqref{eqn:truerisk} in terms of discretized partitions and discretized hypothesis class as follows.
\begin{equation}
\min_{\mathcal{M}\in \mathcal{H}_d^{K}, \mZ \in \mP_K^d(\mY)^{\dagger}} R(f, g_{M, \mZ}; \mD)  \label{eqn:truerisk_d} 
\end{equation}
where  $\mathcal{H}_d$ is the set of discretized constant models.  From the equation \eqref{emse_diff_o}, \eqref{mse_diff} and  \eqref{SE_diff} the minimizer of \eqref{eqn:truerisk_d} will be very close to the minimizer of \eqref{eqn:truerisk}.

We also reformulate the empirical risk minimization \eqref{minim-prob}  in terms of the discrete space below. 

\begin{equation}
\min_{\mathcal{M} \in \mathcal{H}_d^{K},  \mZ \in \mP_K^d(\mY)^{\dagger}}  \hat{R}(\mathcal{M}, \mathcal{Z}; D) 
\label{disc_opt}
\end{equation}

We know that the hypothesis class for the above problem  is finite. We use Corollary 4.6 in \cite{shalev2014understanding} to arrive at the conclusion that empirical risk minimization leads to successful PAC learning of the hypothesis class of all the discrete partitions.  Hence, solving \eqref{disc_opt} (Algorithm 1 and 2's output leads to a solution close to \eqref{disc_opt}) leads to solving \eqref{eqn:truerisk_d}.

 Let us extend the proof to linear hypothesis class.  Each linear function $\beta^tx$ is characterized by a vector $\beta$ and we also assume that $\|\beta\| $ is bounded above by a value $M$.
Suppose we discretize each component of this vector and search for the optimal solution in this discretized space.  The vector $\beta$ has a  bounded norm $\implies$ the discretized space of linear functions consists is a finite set.
The MSE loss $MSE_c$ is minimized when $\bar{y}_k(x) =(\beta^{*})^tx$ and we write the loss w.r.t. $\bar{y}_k(x)$.  The discretization of $\beta^{*}$ is given as $[\beta^{*}]_d$. Therefore, $\bar{y}^d_k(x) =[\beta^{*}]_d^tx$  
$$MSE_{c,l}^{*} =\frac{1}{|D|}\sum_{k=1}^{K} \sum_{x \in D_k}(y-\bar{y}_k(x)^2 $$

$$MSE_{c,l}^{\dagger} = \frac{1}{|D|}\sum_{k=1}^{K} \sum_{x \in D_k}(y-\bar{y}^d_k(x))^2$$

%
%
%
%
%

The difference between the MSEs is given as follows. 

\begin{equation}
\begin{split}
&|MSE_{c,l}^{\dagger} - MSE_{c,l}^* |=  \frac{1}{|D|}\sum_{k=1}^{K} \sum_{x \in D_k}|(2y-\bar{y}_k(x)-\bar{y}^{d}_k(x))|| \bar{y}_k^d(x)-\bar{y}_k(x)| 
\end{split}
\label{mse_diff1}
\end{equation}

The difference $| \bar{y}_k^d(x)-\bar{y}_k(x)| $ can be made arbitrarily small by making the discretization of each component of $\beta$ small.  The rest of the proof follows the exact same steps as in the case of constant hypothesis class as shown above.




\section*{Proof of Proposition 3}
For this proposition, we assume that the loss function is MSE and the interpretive model belongs to the piecewise constant class.
In this proposition, we need to show that the complexity of the Algorithm 1 and 2 is $\mathcal{O}(|D|^3K)$. 

From Algorithm 1 we know that  the main step that is executed inside the for loops is
$$V^{'}(n+1,k) = \min_{n^{'} \in \{1,..,n-1\}}\{V^{'}(n^{'},k-1)+ G(n^{'}+1,n) \}$$

Let us compute the complexity of the above step. Note that the terms inside the above expression depends on $V^{'}(n^{'},k-1)$ and $G(n^{'}+1,n)$. $V^{'}(n^{'},k-1)$ is stored already from the previous $n$ iterations. The computation of $G(n^{'}+1,n) $ takes $\mathcal{O}(n)$ steps at most if the loss is MSE. We need to compare $n$ of these values. Therefore, the total time for this step is $\mathcal{O}(n^2)$ steps. For a fixed $n$ the inner for loop will take $\mathcal{O}(n^2K)$ steps. 

We can bound the steps for the outer for loop as $C\sum_{n=1}^{|D|} n^{2}K $ steps, which grows as $\mathcal{O}(|D|^3 K)$ steps.
The complexity of Algorithm 2 is $\mathcal{O}(K)$ (as there are $K$ calls to the function $\Phi$).  

\section*{Proof of Proposition 4}
We already showed that the optimization problems in equations (8) and (11) in the main manuscript are equivalent.

Next we will show that the optimizing among ordered partitions is as good as optimizing among all the partitions.

$$	
\min_{\mathcal{M} \in \mathcal{H}^{K}, \mZ \in \mP_K(\mX)} \hat{R}(\mathcal{M}, \mZ; D) =  \min_{\mathcal{M} \in \mathcal{H}^{K}, \mZ \in \mP_K(\mX)^{\dagger}} \hat{R}(\mathcal{M}, \mZ; D) 
$$

We now state a property that is used to construct the optimal ordered partition that is as good as the optimal partition.

\textbf{Ordering Property:} Consider a partition $\mathcal{Z}$ of the feature space. Consider any two regions of the partition say $A$ and $B$.  We refer to the sets of points in the dataset that belong to $A$ as $\tilde{A}$ and $B$ as $\tilde{B}$. Define $f(\tilde{A}) =\{f(a),\; \forall a \in \tilde{A} \}$. The set of predictions for sets $\tilde{A}$ and $\tilde{B}$ are $f(\tilde{A})$ and $f(\tilde{B})$. 
The sample means for the predictions at the points in $\tilde{A}$ and $\tilde{B}$ are $\bar{M}(\tilde{A})$ and $\bar{M}(\tilde{B})$ respectively. Without loss of generality assume that $\bar{M}(\tilde{A}) < \bar{M}(\tilde{B})$. The property states that for all the points in $\tilde{A}$ their corresponding black-box predictions $f(x) < \frac{\bar{M}(\tilde{A}) + \bar{M}(\tilde{B})}{2}$ and for all the points in $\tilde{B}$ their corresponding black-box predictions 
$f(x) > \frac{\bar{M}(\tilde{A}) + \bar{M}(\tilde{B})}{2}$. 
If this property holds for every pair of regions in the partition, then it automatically implies that the partition is ordered.

\textbf{Idea.} We will show that if a partition does not satisfy the ordering property then it can always be modified to construct a partition that is ordered and is at least as good as the partition that we start with.


We start with the partition $\mathcal{Z}$  (we are interested in partitions with atleast two regions in them.)  that is optimal. Suppose that $\mathcal{Z}$  does not satisfy the ordering property.

If the ordering property is not satisfied, then there are two possibilities:
\begin{enumerate}
	\item For some two regions $A$ and $B$ in the partition (and corresponding induced sets $\tilde{A}$ and $\tilde{B}$) there exists a point $x_s\in \tilde{A}$  such that $f(x_s) > \frac{\bar{M}(\tilde{A})  + \bar{M}(\tilde{B})}{2} $
	\item For some two regions $A$ and $B$ in the partition (and corresponding induced sets $\tilde{A}$ and $\tilde{B}$) there exists a point $x_s \in \tilde{B}$ such that  $f(x_s) < \frac{\bar{M}(\tilde{A})  + \bar{M}(\tilde{B})}{2} $
\end{enumerate}

For the rest of the proof we will assume that the first case is true. The analysis for the second case would be similar as well.  We will show that we can modify the partition $\mathcal{Z}$ to $\mathcal{Z}^{'}$ in a simple way such that the MSE for $\mathcal{Z}^{'}$ is infact less than or equal to the MSE of $\mathcal{Z}$. 

We modify the set $\tilde{B}$  by adding $x_s$ to it from the region $\tilde{A}$. We call these new regions as  $\tilde{B}^{'}$ and $\tilde{A}^{'}$  respectively.   We express the difference between the losses before and after moving the $x_s$ below. Let $y_s = f(x_s)$. 

\begin{equation} 
\begin{split}
L_{diff} &=  \sum_{y\in f(\tilde{A})} \big[y-\bar{M}(\tilde{A}) \big]^2 +
\sum_{y\in f(\tilde{B})} \big[y-\bar{M}(\tilde{B}) \big]^2  -\sum_{y\in f(\tilde{A}^{'})} \big[y-\bar{M}(\tilde{A}^{'})\big]^2  
- \sum_{y\in f(\tilde{B}^{'})} \big[y-\bar{M}(\tilde{B}^{'})\big]^2 \\
=& \big( |\tilde{A}| - 1\big)\bar{M}^{2}(\tilde{A}^{'}) + \big( |\tilde{B}| + 1\big)\bar{M}^2(\tilde{B}^{'}) -  |\tilde{A}| \bar{M}^2(\tilde{A})- |\tilde{B}| \bar{M}^2(\tilde{B})
\end{split}
\label{Ldiff}
\end{equation}

Our objective is to show that $L_{diff} \geq 0$.


We express $\bar{M}(\tilde{A}^{'})$ and $\bar{M}(\tilde{B}^{'})$ in terms of $\bar{M}(\tilde{A})$ and $\bar{M}(\tilde{B})$ respectively as follows. 


\begin{equation}
\begin{split}
\bar{M}(\tilde{A}^{'}) =  \frac{\bar{M}(\tilde{A}) |\tilde{A}| - y_s}{|\tilde{A}|-1} \\ 
\bar{M}(\tilde{B}^{'}) =  \frac{\bar{M}(\tilde{B}) |\tilde{B}| + y_s}{|\tilde{B}|+1} \\ 
\end{split}
\end{equation}

\begin{equation}
\begin{split}
\bar{M}(\tilde{A}^{'})^2 =  \big(\frac{\bar{M}(\tilde{A}) |\tilde{A}| - y_s}{|\tilde{A}|-1} \big)^2 =   \frac{\bar{M}(\tilde{A})^2|\tilde{A}|^2  + y_s^2 -2y_s \bar{M}(\tilde{A})  }{(|\tilde{A}|-1)^2} 
\end{split}
\label{MA}
\end{equation}

\begin{equation}
\begin{split}
\bar{M}(\tilde{B}^{'})^2 =  \big(\frac{\bar{M}(\tilde{B}) |\tilde{B}| + y_s}{|\tilde{B}|+1} \big)^2 =  \frac{\bar{M}(\tilde{B})^2|\tilde{B}|^2  + y_s^2 +2y_s \bar{M}(\tilde{B})  }{(|\tilde{B}|+1)^2} 
\end{split}
\label{MB}
\end{equation}

We substitute \eqref{MA} and \eqref{MB} into \eqref{Ldiff} to obtain the following.

\begin{equation}
\begin{split}
L_{diff} =   \frac{\bar{M}^{2}(\tilde{A})|\tilde{A}|  + y_s^2 -2y_s \bar{M}(\tilde{A}) |\tilde{A}| }{(|\tilde{A}|-1)} + \frac{-\bar{M}^{2}(\tilde{B})|\tilde{B}|  + y_s^2 +2y_s \bar{M}(\tilde{B}) |\tilde{B}|}{(|\tilde{B}|+1)}  
\end{split}
\label{Ldiff_s}
\end{equation}

The expression in the  above  equation \eqref{Ldiff_s} is a quadratic function of $y_s$. We call it $L_{diff}(y_s)$.  We want to analyze the behavior of the above function in the region $y_s > \frac{\bar{M}(\tilde{A}) + \bar{M}(\tilde{B})}{2}$. Our objective is to show that the above function is greater than zero when $y_s > \frac{\bar{M}(\tilde{A}) + \bar{M}(\tilde{B})}{2}$.
We compute the gradient of the above function at $y_s  = \frac{\bar{M}(\tilde{A}) + \bar{M}(\tilde{B})}{2}$ as \eqref{Ldiff_der}.

\begin{equation}
\begin{split}
& \frac{d L_{diff}(y_s)}{ dy_s} |_{y_{s} =	\frac{\bar{M}(\tilde{A}) + \bar{M}(\tilde{B})}{2}} =\Big[\frac{2y_s}{|\tilde{A}|-1} + \frac{2y_s}{|\tilde{B}|+1} -2 \bar{M}(\tilde{A}) \frac{|\tilde{A}|}{|\tilde{A}|-1} + 2 \bar{M}(\tilde{B})\frac{|\tilde{B}|}{|\tilde{B}|+1}\Big]_{y_s=\frac{\bar{M}(\tilde{A}) + \bar{M}(\tilde{B})}{2}} \\
&=\Big[2 y_s(|\tilde{A}|+|\tilde{B}|)\frac{1}{(|\tilde{A}|-1)(|\tilde{B}|+1)} -2 \bar{M}(\tilde{A}) \frac{|\tilde{A}|}{|\tilde{A}|-1} + 2 \bar{M}(\tilde{B})\frac{|\tilde{B}|}{|\tilde{B}|+1}\Big]_{y_s=\frac{\bar{M}(\tilde{A}) + \bar{M}(\tilde{B})}{2}} \\
&=\Big[ \frac{2y_s(|\tilde{A}|+|\tilde{B}|)+ 2|\tilde{A}||\tilde{B}|(\bar{M}(\tilde{B})- \bar{M}(\tilde{A})) - 2\bar{M}(\tilde{A}) |\tilde{A}| -2\bar{M}(\tilde{B}) |\tilde{B}|}{(|\tilde{A}|-1)(|\tilde{B}|+1)}\Big]_{y_s = \frac{\bar{M}(\tilde{A})+ \bar{M}(\tilde{B})}{2}} \\
& =\bar{M}(\tilde{A})(|\tilde{B}|-|\tilde{A}|)  + \bar{M}(\tilde{B})(|\tilde{A}| -|\tilde{B}|) + 2|\tilde{A}||\tilde{B}|(\bar{M}(\tilde{B})- \bar{M}(\tilde{A})) \\
&=  (|\tilde{A}| -|\tilde{B}|) (\bar{M}(\tilde{B}) - \bar{M}(\tilde{A}))+ 2|\tilde{A}||\tilde{B}|(\bar{M}(\tilde{B})- \bar{M}(\tilde{A})) \\ 
&=  (\bar{M}(\tilde{B})- \bar{M}(\tilde{A})) (|\tilde{A}| -|\tilde{B}| + 2|\tilde{A}||\tilde{B}|) \\ 
&= (\bar{M}(\tilde{B})- \bar{M}(\tilde{A}))(|\tilde{A}| + |\tilde{B}|(2|\tilde{A}|-1))
\end{split}
\label{Ldiff_der}
\end{equation}

Since $\bar{M}(\tilde{B}) \geq \bar{M}(\tilde{A})$ and $|\tilde{A}|\geq 1$ the  expression in \eqref{Ldiff_der} is greater than zero.  
Note that 
$$\frac{d^2 L_{diff}(y_s)}{ dy_s^2} = \frac{2}{|\tilde{A}|-1} + \frac{2}{|\tilde{B}|+1}\geq 0$$
Therefore,  the gradient of the above expression in \eqref{Ldiff_der} will be greater than zero at all the points greater than $\frac{\bar{M}(\tilde{A}) + \bar{M}(\tilde{B})}{2}$. Hence, we get $$\min_{y_s\in [\frac{\bar{M}(\tilde{A})+ \bar{M}(\tilde{B})}{2}, \infty) } L_{diff}(y_s) = L_{diff}(\frac{\bar{M}(\tilde{A}) + \bar{M}(\tilde{B})}{2})$$ 
Next, we compute $L_{diff}(\frac{\bar{M}(\tilde{A}) + \bar{M}(\tilde{B})}{2})$ in \eqref{sim_diff} and we see that the expression is always greater than or equal to zero. 

\begin{equation}
\begin{split} &L_{diff}(\frac{\bar{M}(\tilde{A}) + \bar{M}(\tilde{B})}{2})=\\ &\Big[\frac{\bar{M}(\tilde{A})^2|\tilde{A}|  + y_s^2 -2y_s \bar{M}(\tilde{A}) |\tilde{A}| }{(|\tilde{A}|-1)} +  \frac{-\bar{M}(\tilde{B})^2|\tilde{B}|  + y_s^2 +2y_s \bar{M}(\tilde{B})  |\tilde{B}|}{(|\tilde{B}|+1)}\Big]_{y_s = \frac{\bar{M}(\tilde{A})+ \bar{M}(\tilde{B})}{2}}  \\&= \frac{\bar{M}^2(\tilde{A})|\tilde{A}|  +[\frac{\bar{M}(\tilde{A}) + \bar{M}(\tilde{B})}{2}]^2 -[\bar{M}(\tilde{A}) + \bar{M}(\tilde{B})] \bar{M}(\tilde{A}) |\tilde{A}| }{(|\tilde{A}|-1)} + \\ & \frac{-\bar{M}(\tilde{B})^2|\tilde{B}|  + [\frac{\bar{M}(\tilde{A}) + \bar{M}(\tilde{B})}{2}]^2 +[\bar{M}(\tilde{A}) + \bar{M}(\tilde{B})]  \bar{M}(\tilde{B}) |\tilde{B}|}{(|\tilde{B}|+1)}  \\& = 
\frac{[\frac{\bar{M}(\tilde{A}) + \bar{M}(\tilde{B})}{2}]^2 - \bar{M}(\tilde{B}) \bar{M}(\tilde{A}) |\tilde{A}| }{(|\tilde{A}|-1)} + \frac{ [\frac{\bar{M}(\tilde{A}) + \bar{M}(\tilde{B})}{2}]^2 +\bar{M}(\tilde{A})  \bar{M}(\tilde{B})  |\tilde{B}|}{(|\tilde{B}|+1)} \\&=
\frac{[\bar{M}(\tilde{A}) + \bar{M}(\tilde{B})]^2}{4} \frac{|\tilde{B}|+|\tilde{A}|}{(|\tilde{A}|-1)(|\tilde{B}|+1)} -\bar{M}(\tilde{A})  \bar{M}(\tilde{B}) \frac{|\tilde{A}|+ |\tilde{B}|}{(|\tilde{A}|-1)(|\tilde{B}|+1)}  \\ &=\frac{|\tilde{B}|+|\tilde{A}|}{(|\tilde{A}|-1)(|\tilde{B}|+1)} [\frac{[\bar{M}(\tilde{A}) + \bar{M}(\tilde{B})]^2}{4}  -\bar{M}(\tilde{A})  \bar{M}(\tilde{B}) ] \\&=\frac{|\tilde{B}|+|\tilde{A}|}{(|\tilde{A}|-1)(|\tilde{B}|+1)} [\bar{M}(\tilde{B}) - \bar{M}(\tilde{A})]^2/4 
\end{split}
\label{sim_diff}
\end{equation}

If $\bar{M}(\tilde{B}) >  \bar{M}(\tilde{A})$, then this contradicts the optimality of the partition $\mathcal{Z}$. If $\bar{M}(\tilde{B}) = \bar{M}(\tilde{A})$, then the partition $\mathcal{Z}^{'}$ is as good as $\mathcal{Z}$. The partition $\mathcal{Z}^{'}$ may not be ordered. We can repeat the above argument starting with $\mathcal{Z}^{'}$ until we have an ordered partition that has at least the same loss as $\mathcal{Z}$.  Note that $\tilde{A}$ has to have at least two points for the setup to make sense.  If $\tilde{A}$ only had one point then shifting the point would reduce the number of regions in the partition.   In the case when $|\tilde{A}|$, we do not shift the point from $\tilde{A}$ to $\tilde{B}$ but instead we swap the point $y_s$ with a point from $\tilde{B}$ which has a lower value than $y_s$.  The same conclusion follows for this case as well.

\section*{Generalization of k-means clustering}

In the next propositon, we discuss the extension of Proposition 4 to a  general loss function $\ell$. We assume $\ell(|.|)$ is strictly convex (for e.g., $\ell(|x|) =x^4$ ).

\textbf{Proposition 5.} If Assumption 3 and 4 hold, and $\ell(|.|)$ is strictly convex and differentiable almost everywhere, then	for every $\epsilon, \delta > 0$ and every $K$ there is some $m^*(\epsilon, \delta, K)$ such that if the training set $D$ is drawn i.i.d. from the distribution 
$\mathcal D$ and $|D| \geq m^*(\epsilon, \delta, K)$, then with probability at least $1 - \delta$ we have $
\bigl| 	\min_{\mathcal{M} \in \mathcal{H}^{K},  \mathcal{ I} \in \mP_K(\mY)}  \hat{R}(\mathcal{M}, \mathcal{I}; D)  -  \hat{R}(\mathcal{M}^{\#},\mZ^{\#}; D) \big| < \epsilon$

\textbf{Proof.}
From Proposition 1, we know that $\hat{R}(\mathcal{M}^{\#},\mZ^{\#}; D) =\min_{\mathcal{M} \in \mathcal{H}^{K}, \mZ \in \mP_K(\mX)^{\dagger}} \hat{R}(\mathcal{M}, \mZ; D) $

We need to show that	$$	
|\min_{\mathcal{M} \in \mathcal{H}^{K}, \mZ \in \mP_K(\mX)} \hat{R}(\mathcal{M}, \mZ; D) - \min_{\mathcal{M} \in \mathcal{H}^{K}, \mZ \in \mP_K(\mX)^{\dagger}} \hat{R}(\mathcal{M}, \mZ; D) | \leq \epsilon 
$$	
We denote $\ell(|x-y|)$ as $L(x,y)$.
The optimal value of the constant cluster mapping only depends on the data points in that cluster/region in the partition and it is computed as follows. For cluster $Z$ we write the local optimal value as $y^{*}_{Z}$. 

\textbf{Dense Partitions:} 
We first show that it is sufficient to consider a certain type of partitions to guarantee approximate optimality, which we call dense partitions. The idea behind a dense partition is described as follows.
As the dataset becomes large, each set in the induced partition should also become large. For ease of explanation, we will use the induced partitions on the dataset only. 

Suppose the total number of data points is $n = |D|$. 

\textbf{Definition.} Consider a partition $\mathcal{Z} = \{Z_1,..., Z_K\}$. The set of points that belong to $Z_j$ are given as $\tilde{Z}_j$ and let $n_j = |\tilde{Z}_j|$. We refer to $\mathcal{Z}$ as dense if the data size grows to infinity, then the size of each induced region should also increase to infinity, i.e., as $n \rightarrow \infty \implies $ $n_j \rightarrow \infty$, $\forall j \in \{1,...,K\}$.

Let $\mathcal{P}_{K}^{d}(\mathcal{X})$ be the set of all the dense partitions of $\mathcal{X}$. We argue that it is sufficient to search in the space of dense partitions provided the dataset is large enough. Suppose that there is a partition $\mathcal{Z} \in \mathcal{P}_{K}^{d}(\mathcal{X})^{c}$, which is not dense. If a partition is not dense, then it can be argued that there exists a certain region $Z_k$ such that $n_k \leq N_k$, where $N_k$ is the upper bound on the size of $Z_k$. We construct a new partition from $\mathcal{Z}$.  We transfer the points in $Z_k$ to one of the remaining regions. The maximum change in the loss can be bounded by $c\frac{N_k}{n}$ for some $c>0$. If the data size is large enough, then the change in loss can be bounded less than $\epsilon/K$. We can repeat this argument for all the regions that have a bounded number of points. The final partition we get as a result will be a dense partition and its loss will be close to the original partition. Therefore, for the rest of the proof we restrict our attention to dense partitions.

\textbf{Ordering property for general loss function:} We now state a property that is very similar to the property (basically a generalization) we stated for MSE, used to construct the optimal ordered partition. Consider a partition $\mathcal{Z}$. Consider any two regions of the partition say $A$ and $B$ (with induced sets on the data given as $\tilde{A}$  and $\tilde{B}$) and the corresponding optimal predicted values assigned by the model $M$ are $y^{*}_{\tilde{A}}$ and $y^{*}_{\tilde{B}}$ respectively. Without loss of generality assume that $y^{*}_{\tilde{A}} < y^{*}_{\tilde{B}} $. The property states that for all the points in $\tilde{A}$ their corresponding black-box predictions $f(x) < \frac{y^{*}_{\tilde{A}}+ y^{*}_{\tilde{B}}}{2}$ and for all the points in $\tilde{B}$ their corresponding black-box predictions 
$f(x) > \frac{y^{*}_{\tilde{A}}+ y^{*}_{\tilde{B}}}{2}$. Note that if this property holds for every pair of regions in the partition, then it automatically implies that the partition is ordered.

We start with the partition $\mathcal{Z}$  (we are interested in partitions with   at least two regions in them)  that is optimal and does not satisfy the ordering property. 

There are two possibilities:
\begin{enumerate}
	\item For some two regions $\tilde{A}$ and $\tilde{B}$ in the partition there exists a point $x_s\in \tilde{A}$  such that $f(x_s) > \frac{y^{*}_{\tilde{A}}+ y^{*}_{\tilde{B}}}{2}$
	\item For some two regions $\tilde{A}$ and $\tilde{B}$ in the partition there exists a point $x_s \in \tilde{B}$ such that  $f(x_s) < \frac{y^{*}_{\tilde{A}}+ y^{*}_{\tilde{B}}}{2} $
\end{enumerate}
For the rest of the proof we will assume that the first case is true. The analysis for the second case would be similar as well.  



\textbf{Idea.1} We will show that we can modify the partition $\mathcal{Z}$ to $\mathcal{Z}^{'}$ in a simple way such that the loss for $\mathcal{Z}^{'}$ is infact lower than the loss  of $\mathcal{Z}$. We modify the region $\tilde{B}$ of  by adding $x_s$ to it from the region $\tilde{A}$. We call these new regions as $\tilde{B}^{'}$ and $\tilde{A}^{'}$ respectively. 

\textbf{Idea 2.} In this case since we deal with general loss functions the optimal value $y^{*}_{\tilde{A}}$ does not have a closed form unlike the case of MSE. This makes it difficult to track the change in $y^{*}_{\tilde{A}}$ when we construct the new regions $\tilde{A}^{'}$. However, we can track the changes using influence functions \cite{cook1982residuals}  provided the number of data points is sufficiently large.

We express the difference between the losses before  and after moving the $x_s$. Let $y_s = f(x_s)$  Since  $y_s >\frac{y^{*}_{\tilde{A}}+ y^{*}_{\tilde{B}}}{2} $ we get 
$L(y_s,y_{\tilde{A}}^{*})>L(y_s,y_{\tilde{B}}^{*})$.  

We define the loss for the sets $\tilde{A}$ and $\tilde{B}$ as 

$L^{total}(\tilde{A}) = \sum_{y \in \tilde{A}} L(y, y_{\tilde{A}}^{*}) $

$L^{total}(\tilde{B}) = \sum_{y \in \tilde{B}} L(y, y_{\tilde{B}}^{*}) $


We write the change in the loss function for the two sets $A$ and $B$ as follows.  

\begin{equation}
\begin{split}
&L^{total}(\tilde{A}) - L^{total}(\tilde{A}^{'}) =\sum_{y \in \tilde{A}, y\not=y_s} L(y, y_{\tilde{A}}^{*}) - \sum_{y \in \tilde{A}, y\not=y_s}L(y, y_{\tilde{A}^{'}}^{*}) + L(y_s, y_{\tilde{A}}^{*})
\end{split}
\label{totala}
\end{equation} 

\begin{equation}
\begin{split}
&L^{total}(\tilde{B}) - L^{total}(\tilde{B}^{'})=\sum_{y \in \tilde{B}, y\not=y_s} L(y, y_{\tilde{B}}^{*}) - \sum_{y \in \tilde{B}, y\not=y_s}L(y, y_{\tilde{B}^{'}}^{*}) - L(y_s, y_{\tilde{B}^{'}}^{*})
\end{split}
\label{totalb}
\end{equation}

We track the change in $y_{\tilde{A}}^{*}$ to $y_{\tilde{A}^{'}}^{*}$ and $y_{\tilde{B}}^{*}$ to $y_{\tilde{B}^{'}}^{*}$  using influence functions  \cite{cook1982residuals}.  

We can express the difference $y_{\tilde{A}^{'}}^{*} -y_{\tilde{A}}^{*}$ using \cite{cook1982residuals}.
$$y_{\tilde{A}^{'}}^{*} -y_{\tilde{A}}^{*} \approx \Big[\frac{\partial L(y, c) }{\partial c}|_{c=y_{\tilde{A}}^{*}}\Big] \frac{1}{\sum_{y \in \tilde{A}}\frac{\partial^2L(y,c)}{\partial c^2}|_{c=y_{\tilde{A}}^{*}}}\frac{1}{|\tilde{A}|}$$

\begin{equation}
\begin{split}
& L(y ,y_{\tilde{A}}^{*}) - L(y, y_{\tilde{A}^{'}}^{*}) \approx \frac{\partial L(y, c) }{\partial c}|_{c=y_{\tilde{A}}^{*}}  (y_{\tilde{A}}^* - y_{\tilde{A}^{'}}^{*})
\end{split}
\label{eqn-diff1}
\end{equation}

We substitute $y_{\tilde{A}^{'}}^{*} -y_{\tilde{A}}^{*}$ from above in \eqref{eqn-diff1}.

\begin{equation}
L(y ,y_{\tilde{A}}^{*}) - L(y, y_{\tilde{A}^{'}}^{*}) \approx -\Big[\frac{\partial L(y, c) }{\partial c}|_{c=y_{\tilde{A}}^{*}}\Big]^2 \frac{1}{\sum_{y \in \tilde{A}}\frac{\partial^2L(y,c)}{\partial c^2}|_{c= y_{\tilde{A}}^{*}}} \frac{1}{|\tilde{A}|}
\end{equation}


We track the change in $y_{\tilde{B}}^{*}$ to $y_{\tilde{B}^{'}}^{*}$.
The final expressions for the change are given as follows. 

\begin{equation}
\begin{split}
&L(y ,y_{\tilde{A}}^{*}) - L(y, y_{\tilde{A}^{'}}^{*})\approx -\Big[\frac{\partial L(y, c) }{\partial c}|_{c=y_{\tilde{A}}^{*}}\Big]^2 \frac{1}{\sum_{y \in \tilde{A}}\frac{\partial^2L(y,c)}{\partial c^2}|_{c= y_{\tilde{A}}^{*}}} \frac{1}{|\tilde{A}|}
\end{split}
\label{inf1}
\end{equation}

\begin{equation}
\begin{split}
& L(y ,y_{\tilde{B}}^{*}) - L(y, y_{\tilde{B}^{'}}^{*}) \approx  \Big[\frac{\partial L(y, c) }{\partial c}|_{c=y_{\tilde{B}}^{*}}\Big]^2 \frac{1}{\sum_{y \in \tilde{B}}\frac{\partial^2L(y,c)}{\partial c^2}|_{c= y_{\tilde{B}}^{*}}} \frac{1}{|\tilde{B}|}
\end{split}
\label{inf2}
\end{equation}

The function  $\frac{\partial L(y, c) }{\partial c} $ is continuous almost everywhere. The set $\mathcal{Y}$ is an interval. Hence, the above function $\frac{\partial L(y, c) }{\partial c} $ on the interval $\mathcal{Y}$  is  bounded. 

Consider $\sum_{y \in \tilde{B}}\frac{\partial ^2L(y,c)}{\partial c^2}|_{c= y_{\tilde{B}}^{*}}$. We claim that if $|\tilde{B}|$ is sufficiently large, then $$\sum_{y \in \tilde{B}}\frac{\partial^2L(y,c)}{\partial c^2}|_{c= y_{\tilde{B}}^{*}} =  \Omega(|\tilde{B}|), \;\; \text{with a high probability} $$
i.e. there exists a $k$ and $n_0$ such that $\forall \;|\tilde{B}| > n_0$, $\sum_{y \in \tilde{B}}\frac{\partial^2L(y,c)}{\partial c^2}|_{c= y_{\tilde{B}}^{*}}  \geq k |\tilde{B}|$ with a high probability. We know that $L$ is strictly convex.  The term $\frac{1}{|\tilde{B}|}\sum_{y \in \tilde{B}}\frac{\partial^2L(y,c)}{\partial c^2}|_{c= y_{\tilde{B}}^{*}} $  will take a fixed positive value in limit as $\tilde{B}$ grows large (from strong law of large numbers).  We can set $k$ to be anything smaller than the limit of this term to establish the claim. Note that since we are using dense partitions it is safe to assume that as the data will grow large so will the size of the regions.
Similarly $\sum_{y \in \tilde{A}}\frac{\partial^2L(y,c)}{\partial c^2}|_{c= y_{\tilde{A}}^{*}} =  \Omega(|\tilde{A}|)$.

Therefore, we can substitute the lower bounds on  $\sum_{y \in \tilde{B}}\frac{\partial^2L(y,c)}{\partial c^2}|_{c= y_{\tilde{B}}^{*}}$ and $\sum_{y \in \tilde{A}}\frac{\partial^2L(y,c)}{\partial c^2}|_{c= y_{\tilde{A}}^{*}} $ in \eqref{inf1} and \eqref{inf2} to obtain
\begin{equation}
\begin{split}
& | L(y ,y_{\tilde{A}}^{*}) - L(y, y_{\tilde{A}^{'}}^{*})| \leq \frac{L_{\tilde{A}}}{|\tilde{A}|^2}
\end{split}
\label{ibd1}
\end{equation}
\begin{equation}
\begin{split}
& | L(y ,y_{\tilde{B}}^{*}) - L(y, y_{\tilde{B}^{'}}^{*})| \leq \frac{L_{\tilde{B}}}{|\tilde{B}|^2}
\end{split}
\label{ibd2}
\end{equation}

%
We add \eqref{totala} and \eqref{totalb} to obtain $L_{diff}$ given as follows.  In the simplification below we use \eqref{inf1} and \eqref{inf2}. 

\begin{equation}
\begin{split}
&L_{diff} = L^{total}(\tilde{A}) - L^{total}(\tilde{A}^{'}) + L^{total}(\tilde{B}) - L^{total}(\tilde{B}^{'}) =\sum_{y \in \tilde{A}, y\not=y_s} L(y, y_{\tilde{A}}^{*}) - \sum_{y \in \tilde{A}, y\not=y_s}L(y, y_{\tilde{A}^{'}}^{*}) \\
& + L(y_s, y_{\tilde{A}}^{*})  +
\sum_{y \in \tilde{B}, y\not=y_s} L(y, y_{\tilde{B}}^{*}) -\sum_{y \in \tilde{B}, y\not=y_s}L(y, y_{\tilde{B}^{'}}^{*}) - L(y_s, y_{\tilde{B}^{'}}^{*}) \\ 
& \approx \Big[L(y_s,y_A^{*})- L(y_s, y_B^{*})\Big]  - \sum_{y \in \tilde{A}, y \not=y_s}\Big[\frac{\partial L(y, c) }{\partial c}|_{c=y_{\tilde{A}}^{*}}\Big]^2 \frac{1}{\sum_{y \in \tilde{A}}\frac{\partial ^2L(y,c)}{\partial c^2}|_{c= y_{A}^{*}}} \frac{1}{|\tilde{A}|}\\ 
&  + \sum_{y \in \tilde{B}, y\not=y_s}\Big[\frac{\partial L(y, c) }{\partial c}|_{c=y_{\tilde{B}}^{*}}\Big]^2 \frac{1}{\sum_{y \in \tilde{B}}\frac{\partial^2L(y,c)}{\partial c^2}|_{c= y_{\tilde{B}}^{*}}} \frac{1}{|\tilde{B}|} \\ 
\end{split}
\end{equation} 

The first term in the above expression, i.e.,  $\Big[L(y_s,y_{\tilde{A}}^{*})- L(y_s, y_{\tilde{B}}^{*})\Big]>0$. Therefore, $\exists\; \epsilon^{'}>0$ such that 
\begin{equation}
\Big[L(y_s,y_{\tilde{A}}^{*})- L(y_s, y_{\tilde{B}}^{*})\Big]\geq \epsilon^{'}
\label{epsp}
\end{equation}

The rest of the terms in the above expression are bounded as well. We use the expressions in \eqref{ibd1} and \eqref{ibd2} to arrive at a lower bound on $L_{diff}$ given as 

\begin{equation}
L_{diff} \geq \Big[L(y_s,y_{\tilde{A}}^{*})- L(y_s, y_{\tilde{B}}^{*})\Big]  -\frac{L_{\tilde{A}}}{|\tilde{A}|} - \frac{L_{B}}{|\tilde{B}|}
\end{equation}
Suppose $\frac{L_{\tilde{A}}}{|\tilde{A}|}>\frac{L_{\tilde{B}}}{|\tilde{B}|}$ without loss of generality.
\begin{equation}
L_{diff} \geq \Big[L(y_s,y_{\tilde{A}}^{*})- L(y_s, y_{\tilde{B}}^{*})\Big]  -2\frac{L_{\tilde{A}}}{|\tilde{A}|} 
\end{equation}

Also, from \eqref{epsp} we have
\begin{equation}
L_{diff} \geq \epsilon^{'}  -2\frac{L_{\tilde{A}}}{|\tilde{A}|} 
\end{equation}
Suppose $|\tilde{A}|\geq 4\frac{1}{\epsilon^{'}}{L_{\tilde{A}}}$. Then $L_{diff} \geq \frac{\epsilon^{'}}{2}$. (Note that we are only considering dense partitions. If the data set is large enough the size of $\tilde{A}$ will satisfy the required assumption.)

%
%
%
%

Therefore, $L_{diff}>0$, which means that the loss improves by shifting the data points. This contradicts the optimality of the partition among the dense partitions.


\begin{algorithm*}[tbh]
	\caption{ Approximate computation value and index functions}
	\label{alg:example2}
	\begin{algorithmic}[1]
		\State \textbf{Input:} Dataset $D$, Number of intervals $H$ and number of regions to divide each interval $W$, $\delta$ is an integer, larger the value of $\delta$ the higher the approximation factor
		\State\textbf{Initialize:} Define $V^{'}(1,k)=0, \forall k \in \{1,...,K\}$. 
		\State  For each  $x_i \in D, x_j \in D$ such that $i \leq j$, define $D(i,j) = \{x : x\in D  \; \text{and}\;  \|f(x_i)\| \leq \|f(x)\| \leq \|f(x_j)\|\}$
		\State $ \{S_{1}^{ij},..,S_{W}^{ij}\}= \mathsf{Kmeans}(D(h_l, h_u))$
		\State  $G(i,j) = \sum_{u}\min_{h \in \mathcal{H}} \sum_{ x_r \in S^{ij}_{u}} l(|f(x_r)-h(x_r)|)   $ 
		\State  $M(S_{u}^{ij}) = \arg\min_{h \in \mathcal{H}} \sum_{ x_r \in S^{ij}_{u}} l(|f(x_r) -h(x_r)|)$ 		
		\For{ $n \in \{2,..., |D|\}$}
		\For{$k \in \{1,..., K\}$}
		\State $\;\;\;\;\;\;\;\;\;\;$ \begin{equation} V^{'}(n,k) = \min_{n^{'}\in \{1,\delta+1, 2\delta+1,..,\floor*{\frac{n-1}{\delta}}\delta \}} \big[V^{'}(n^{'},k-1) + G(n^{'}+1, n)\big] \end{equation} 
		\State  $\;\;\;\;\;\;\;\;\;\;$ \begin{equation} \Phi(n,k) = \argmin_{n^{'}\in \{1,\delta+1, 2\delta+1,..,\floor*{\frac{n-1}{\delta}}\delta} \big[V^{'}(n^{'},k-1) + G(n^{'}+1, n)\big] \end{equation}
		\EndFor
		\EndFor
		\State  \textbf{Output:} Value function $V^{'}$, Index function $\Phi$			
	\end{algorithmic}
\end{algorithm*}

\bibliographystyle{IEEEtran}
\bibliography{Piecewise_approximation_summ}

\end{document}